%% file: miedb_nips2026.tex
\documentclass{article}

\ifdefined\nipsanonymous
    \usepackage[eandd]{neurips_2026}
\else
    \usepackage[preprint]{neurips_2026}
\fi

\usepackage[utf8]{inputenc} 
\usepackage[T1]{fontenc}    
\usepackage{hyperref}       
\usepackage{url}            
\usepackage{booktabs}       
\usepackage{amsfonts}       
\usepackage{nicefrac}       
\usepackage{microtype}      
\usepackage{xcolor}         

\usepackage{graphicx}
\usepackage{subcaption}

\usepackage{amsmath}
\usepackage{amssymb}
\usepackage{mathtools}
\usepackage{amsthm}

\usepackage{multirow}
\usepackage{makecell}
\usepackage[most]{tcolorbox}
\usepackage{enumitem}

\usepackage{pifont}   
\usepackage{etoc}

\usepackage{xurl}

\usepackage{wrapfig}

\title{MieDB-100k: A Comprehensive Dataset for \\Medical Image Editing}

\author{%
Yongfan Lai$^{1\ 2\ 3\ 4\ *}$ \quad Wen Qian$^{4\ 5\ 6\ *}$ \quad Bo Liu$^{1\ 2}$ \quad Hongyan Li$^{1\ 2}$ \\ 
\quad \textbf{Hao Luo}$^{4\ 5\ 6\ \dagger}$  \textbf{Fan Wang}$^4$ \quad \textbf{Bohan Zhuang}$^{4\ 6}$ 
\quad \textbf{Shenda Hong}$^{3,\dagger}$\\
$^1$State Key Laboratory of General Artificial Intelligence, Beijing, China \\
\quad $^2$School of Intelligence Science and Technology, Peking University, Beijing, China\\
\quad $^3$National Institute of Health Data Science, Peking University, Beijing, China\\
\quad $^4$DAMO Academy, Alibaba Group, Zhejiang, China\\
\quad $^5$hupan lab, zhejiang province\\
\quad $^6$Zhejiang University, Zhejiang, China\\
\texttt{\{laiyf,liubo2022\}@stu.pku.edu.cn},
\texttt{\{leehy,hongshenda\}@pku.edu.cn}\\
 {\texttt{\{qianwen.qw,michuan.lh,f.wan\}@alibaba-inc.com}} \\  {\texttt{bohan.zhuang@gmail.com}}
}

\def\datasetname{MieDB-100k}
\def\numberoftarget{63} 
\def\numberofmodalities{10}
\def\numberofdata{104,267} 
\def\numberofbenchdata{3,397} 

\newcommand{\cmark}{\textcolor{green!70!black}{\ding{51}}} 
\newcommand{\xmark}{\textcolor{red!80!black}{\ding{55}}}   

\begin{document}
\etocdepthtag.toc{mainpart}

\maketitle

\begin{abstract}
The scarcity of high-quality data remains a primary bottleneck in adapting multimodal generative models for medical image editing.
Existing medical image editing datasets often suffer from limited diversity, neglect of medical image understanding and inability to balance quality with scalability. 
To address these gaps, we propose \datasetname, a large-scale, high-quality and diverse dataset for text-guided medical image editing.
It categorizes editing tasks into perspectives of Perception, Modification and Transformation, considering both understanding and generation abilities.
We construct \datasetname\ via a data curation pipeline leveraging both modality-specific expert models and rule-based data synthetic methods, followed by rigorous manual inspection to ensure clinical fidelity.
Extensive experiments demonstrate that model trained with \datasetname\ consistently outperforms both open-source and proprietary models while exhibiting strong generalization ability.
We anticipate that this dataset will serve as a cornerstone for future advancements in specialized medical image editing.
\ifdefined\nipsanonymous
Dataset and code are publicly available at \url{https://anonymous.4open.science/r/MieDB-100k-3BB5}
\else
    Dataset and code are publicly available at \url{https://github.com/Raiiyf/MieDB-100k}
\fi
\end{abstract}

\section{Introduction}


Multimodal generative models \citep{bagel, flux1kontext, omnigen2, step1xedit, gemini3pro, gptimage1, omni-diffusion, uniCom} have developed rapidly in recent years. In natural image domains, generative models are not only gradually unifying text-guided generation and editing tasks, but also progressively expanding their capabilities to encompass image modification and image understanding \citep{bagel, metamorph, omni-diffusion, uniCom}. However, in medical image domains, their performance remains conspicuously limited, especially in the area of unified editing tasks \citep{medebench, medgenbench}. We attribute this performance degradation primarily to a fundamental scarcity of specialized medical image-editing data.

While a few contemporary studies have proposed benchmarks or datasets for medical image editing, they remain insufficient in three key aspects: \textbf{(1) limited diversity in medical image modalities.} 
Unlike general computer vision, clinical imaging encompasses diverse modalities with distinct physical and structural foundations. However, existing research and datasets are restricted to a narrow range of imaging modalities \citep{medbanana50k, medebench}, typically the widely available modalities such as Chest X-rays and CTs, which cannot adequately train or evaluate a model’s ability across diverse clinical settings. 

\begin{figure*}
    \centering
    \includegraphics[width=1.0\linewidth]{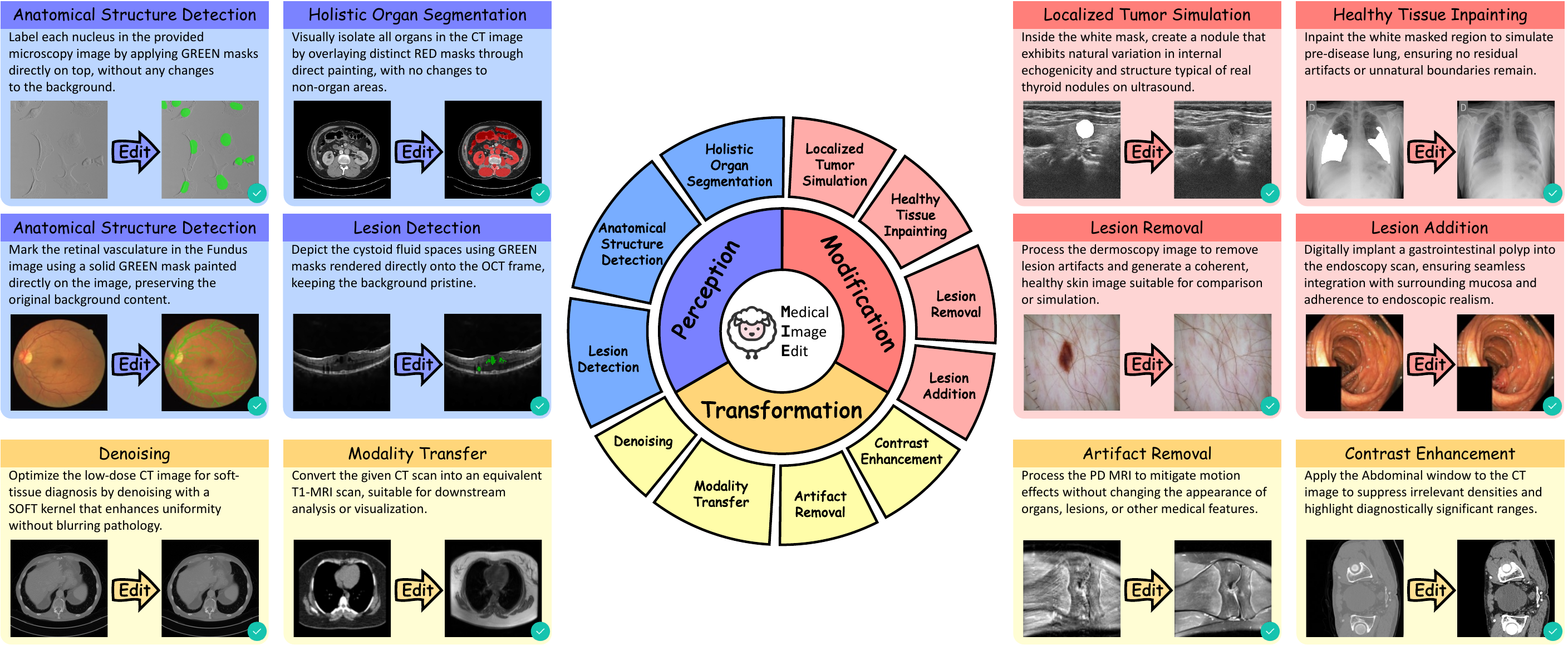}
    \caption{\textbf{\datasetname\ overview.} It categorizes medical image editing tasks into three perspectives, covering diverse medical modalities.}
    \label{fig:overview}
\end{figure*}

\textbf{(2) Neglect of medical image understanding.} 
Almost all medical image editing works focus only on conceptual modification and stylistic transformation tasks, but ignore visual perception tasks (\textit{e.g.}, organ/lesion detection), which has been considered to be beneficial to the generation of image editing models \cite{mingunivision, bagel, meta_unify}.
Additionally, this clinical grounding ensures interpretability and corrects `right-for-the-wrong-reason' edits, which is vital for safety-critical medical applications.
Moreover, recent work \citep{generation_for_understand} has shown that generative models can develop strong visual understanding capabilities, suggesting that neglecting perception tasks may hinder the potential of medical generative models to bridge understanding and generation.

\textbf{(3) Failure to ensure both data quality and scalability.}
Collection of medical image editing data is hindered by the difficulty of generating ground-truth counterfactuals. Some existing studies \citep{medbanana50k, medgenbench} distill general-purpose generative models for quick data scaling. However, these models are not tailored for medical use and hence produce results that lack clinical reliability and explainability. Conversely, previous work \citep{medebench} relies on extensive human involvement to manually collect real medical image pairs, which is notoriously difficult to scale up. 
Moreover, real-world longitudinal data often exhibits spatial misalignment and background inconsistency, as obtaining perfectly calibrated scan pairs is rare in practical medical settings.

In this paper, we address the aforementioned limitations in previous research by introducing \textbf{\datasetname}, a large-scale, high-quality, and diverse dataset for text-guided medical image editing. \textbf{\datasetname} includes \textbf{\numberofdata} editing data, covering \textbf{\numberoftarget} distinct editing targets and \textbf{\numberofmodalities} diverse medical image modalities. We categorize editing tasks into three types: \textbf{Perception}, \textbf{Modification} and \textbf{Transformation}, which consider both model's intrinsic understanding and generation abilities on medical images. To enhance the data fidelity while preserving the scalability, we propose a data curation pipeline leveraging both modality-specific expert models and rule-based data synthetic methods. Additionally, for some complex tasks such as lesion modification, we introduce individuals with medical knowledge to perform manual quality checks on the data to ensure data quality. Finally, we introduced task-specific evaluation metrics to facilitate a comprehensive assessment of the editing models' performance.

We evaluate existing open-source and closed-source multi-modal generative models on \datasetname\ and argue that most of them cannot perform well in medical image editing.
To further validate the reliability and utility of \datasetname, we finetune the OmniGen2 baseline on our dataset. Experimental results demonstrate that \datasetname\ facilitates a substantial performance leap in medical image editing tasks, surpassing or matching SOTA models including Nano Banana Pro. 
It also exhibits strong generalization ability driven by the synergy of understanding and generation tasks.
We anticipate that this dataset will serve as a cornerstone for future advancements in specialized medical image editing.

Our contributions can be summarized as follows:
(1) We propose \textbf{\datasetname}, a large-scale, high-quality and highly diverse dataset for medical image editing with \numberoftarget\ targets and 10 medical image modalities. It is constructed via a credible and scalable data curation pipeline.
(2) We first \textbf{unify the medical image understanding and generation} into the paradigm of edit, and find that joint training yields performance gains for specific tasks. 
(3) We evaluate popular open-source and closed-source multimodal generative models on \textbf{\datasetname}, and observe that training with our data can significantly strengthen the model's capacity for medical image editing.


\section{Related Work}

\subsection{Data Research for Medical Image Editing}

As an emerging area, multimodal medical generative modeling is currently supported by relatively few publicly available datasets for training and benchmarking (Tab.~\ref{tab:benchmark_comparison}). In these works, the primary challenge lies in the construction of high-quality image-edit pairs. MedEBench \citep{medebench}, an early benchmarking effort, curated pairs by manually collecting related images from medical documents. While this ensures clinical validity, the approach lacks scalability. Furthermore, the resulting image pairs often exhibit background inconsistencies, as achieving strict spatial calibration in real-world clinical settings is virtually impossible.
Conversely, Med-banana-50K \citep{medbanana50k} proposed a fully autonomous pipeline where data construction and quality control were managed by Gemini. However, applying general-purpose models to specialized medical scenarios may introduce factual errors or inconsistent edits, raising concerns about data fidelity. Finally, MedGEN-Bench \citep{medgenbench} introduced image-edit pairs using a mix of rule-based and model-based methods; however, the lack of specific architectural details hinders a thorough evaluation of their data quality. 
Moreover, existing benchmarks only focus on content generation evaluation, overlooking the critical aspect of medical image understanding. 
To the best of our knowledge, \datasetname\ is the first work to incorporate medical image editing tasks spanning all three perspectives: Perception, Modification, and Transformation.

\begin{table*}[h]
\centering
\caption{\textbf{Comparison of contemporary medical image editing benchmarks and datasets.} \textbf{P} stands for Perception, \textbf{M} stands for Modification, and \textbf{T} stands for Transformation.}
\resizebox{\linewidth}{!}{
\begin{tabular}{l c c c c c c}
\toprule
\textbf{Benchmark} & \textbf{Size} & \textbf{Modalities} & \textbf{Targets} & \textbf{Perspectives} & \textbf{Source}  & \textbf{Human Inspection} \\ 
\midrule
MedE-Bench \citep{medebench}  & $\sim$1k & 4 & 13 & M & Real & \cmark \\
Med-banana-50K \citep{medbanana50k}  & $\sim$50k & 3 & 23 & M & Synthetic & \xmark  \\
MedGEN-Bench \citep{medgenbench} & $\sim$6k & 6 & 16 & M, T & Real \& Synthetic & \cmark  \\
\midrule
\textbf{MieDB-100k (Ours)} & \textbf{$\sim$100k} & \textbf{\numberofmodalities} & \textbf{\numberoftarget} & \textbf{P, M, T} & \textbf{Real \& Synthetic} & \cmark \\ 
\bottomrule
\end{tabular}
}
\label{tab:benchmark_comparison}
\end{table*}

\subsection{Multimodal Generative Model}

Multimodal generative models \citep{step1xedit, sdxl_turbo, instructpix2pix} accept both images and natural language instructions as input, performing edits by translating semantic commands into precise visual manipulations. Recent studies \citep{qwenimage} often leverage vision-language model encoder and large-scale vision-language pretraining to align the semantic instruction with image modification. For instance, OmniGen2 \citep{omnigen2} utilizes Qwen2.5-VL \citep{qwen25vl} to extract latent representations for semantic alignment, supported by a large-scale, multi-task training strategy.
Furthermore, many recent studies \citep{bagel, omni-diffusion, uniCom} integrate image understanding and editing within a unified architecture. Exploiting these synergies\citep{meta_unify} is essential for creating robust models that are capable of performing both multimodal understanding and visual generation.
On the commercial front, SOTA proprietary models like Gemini-3-Pro-Image (Nano Banana Pro) \citep{gemini3pro} exhibit sophisticated image manipulation abilities, further realizing the real-world potential of multi-modal generative models.
Despite these advancements, current models still struggle with the complexities of medical imaging\citep{medebench, medgenbench}, highlighting the urgent need for comprehensive datasets to accelerate their adaptation to clinical domains.

\section{\datasetname}
\label{sec:method}

This section introduces \datasetname, a high-quality, rigorous, and highly diverse dataset for medical image editing with more than \numberoftarget\ associated medical targets. 
It contains \numberofdata\ image-editing triplets. Figure~\ref{fig:dist}(a) summarizes the distribution of samples across \numberofmodalities\ imaging modalities.

\subsection{Data Definition}
Each entry in \datasetname\ is a triplet $(I, P, O)$, where $I$ is the input medical image, $P$ is the textual prompt that describes edit operation, and $O$ is the target image.

\subsection{Three Perspectives of \datasetname}
\label{sec:pipeline}

\begin{wrapfigure}{r}{0.5\linewidth}
    \centering
    \includegraphics[width=0.95\linewidth]{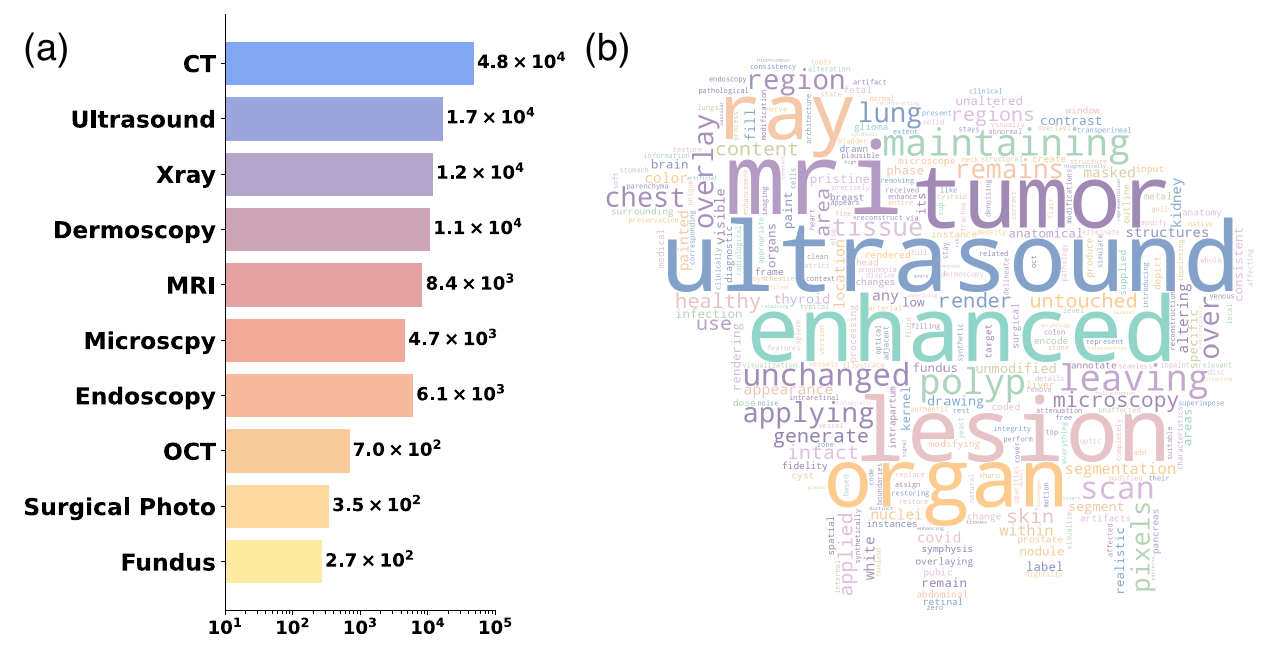}
    \caption{\textbf{Modality distribution (a) and prompt word cloud (b).}}
    \label{fig:dist}
\end{wrapfigure}

\datasetname\ is constructed under a novel categorization of three perspectives, considering both understanding and generation capabilities: (1) \textbf{Perception} tasks, which focus on model's intrinsic medical knowledge via pixel-wise identification of prompted clinical targets in the input image; (2) \textbf{Modification} tasks, which require the model to locate and alter specific medical features; and (3) \textbf{Transformation} tasks, involving medical image restoration, enhancement, and other low-level transformation. 
To ensure the rigor of the data triplets while maintaining scalability, we designed and implemented a specialized data construction pipeline for \datasetname\ (Fig~\ref{fig:pipeline}), and we list all source datasets used for construction in App.~\ref{app:data_list}.

\subsubsection{Perception}
\label{sec:perception}

Perception tasks focus on medical image understanding, and we we formulate it as an editing task by instructing model to generate masks over regions of interest (ROIs), such as specific organs or lesions, through textual prompts. 
Notably, to align with image editing paradigm, the model is prompted to overlay the localization mask directly onto the source image rather than generating a standalone binary mask.
This task serves two primary functions: 
First, since the mask-painting task only requires minimal pixel manipulation (typically modifying a single channel within a specific region), it serves as a \textbf{direct assessment of the medical knowledge embedded in the generative model}, isolating its perceptual accuracy from complex synthesis capabilities.
Second, it introduces a promising application for multimodal generative models in the medical domain: \textbf{assisted interpretation in multimodal manner}. By allowing users to highlight specific targets in medical image through natural language prompts, this approach can assist patients in understanding their diagnostic images, aid medical students in their education, and reduce screening time for senior clinicians. 

\begin{figure*}[t!]
    \centering
    \includegraphics[width=1.0\linewidth]{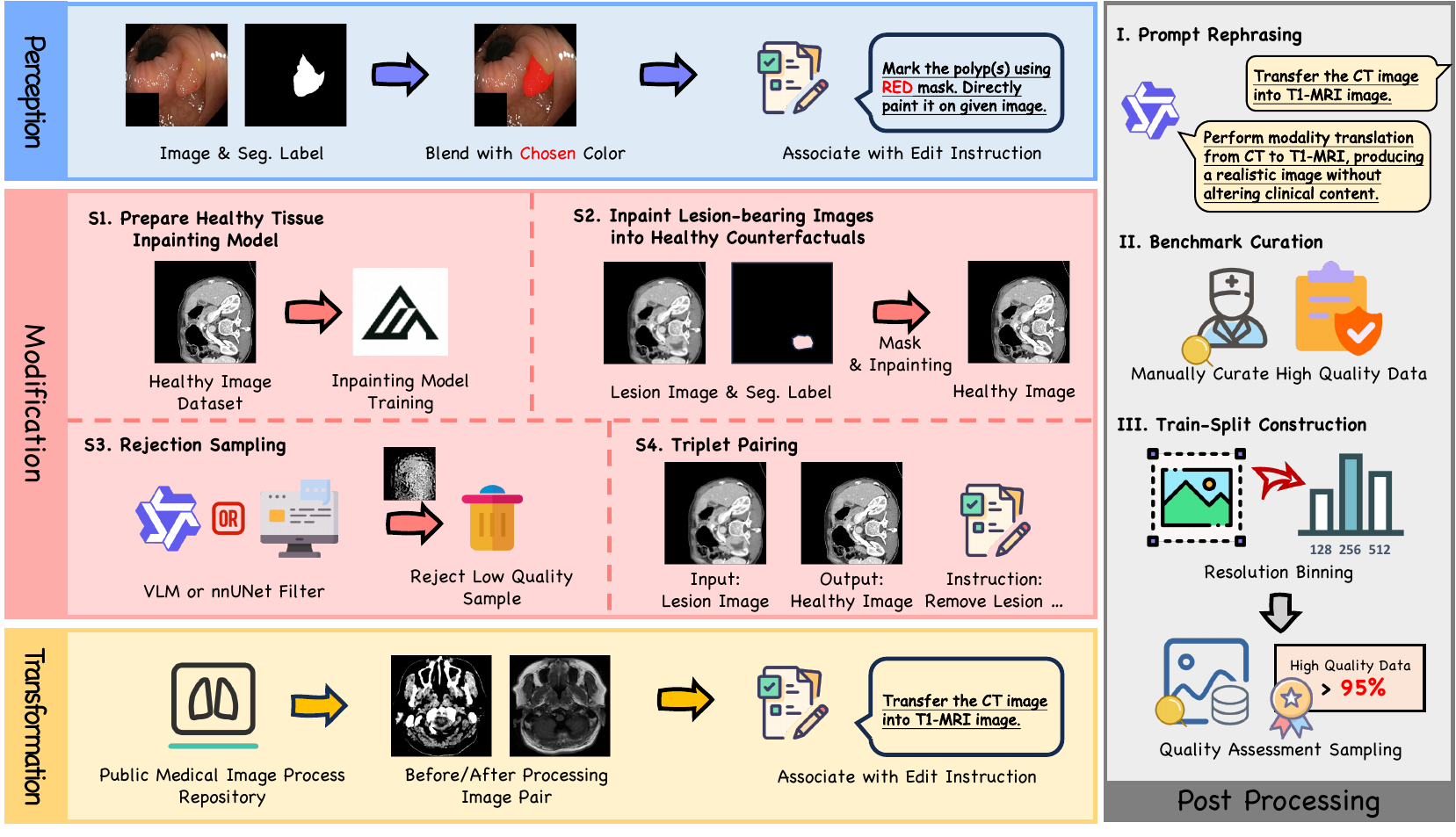}
    \caption{\textbf{Construction pipeline of MieDB-100k.}}
    \label{fig:pipeline}
\end{figure*}

The rule-based construction process for the Perception task’s data triplets is illustrated in Fig.~\ref{fig:pipeline}. 
Specifically, for a segmentation dataset, the original image serves as the input $I$. The output image $O$ is synthesized by overlaying the ground-truth segmentation label, which is rendered in a randomly selected color (red, green, or blue), onto the input image with an alpha-blending transparency of 0.6.
The ROIs of perception can be classified into three types: anatomical structure (organ, organism and so on), lesion area and holistic segmentation (segment all visible and clinically significant structures). We specifies the perception target and visualizing color scheme in the textual prompt $P$. Since this part of the data is constructed following a definite rule, it can be readily scaled up to a diverse set of medical knowledge assessments and to the associated training dataset by leveraging the extensive body of existing medical segmentation research. 
Finally, to ensure a high-quality final benchmark, we manually filtered the initial data pool to remove trivial, redundant, or incorrectly labeled samples.

\subsubsection{Modification}

The perspective of Modification is specifically designed for semantically modifying medical contents, so as to address the diverse requirements of editing beyond just locate them. 
However, constructing modification data triplets is challenging because counterfactual image pairs cannot be captured simultaneously in the real world. 
While one could theoretically leverage general-purpose generative models (\textit{e.g.}, Nano Banana Pro or Qwen-Image-Edit) to produce these edits, such models are not specialized for the medical domain, and therefore are prone to severe hallucinations, which is unacceptable in a healthcare context. 
To construct rigorous edit triplets and preserve scalability, we propose a four-stage process (Fig.~\ref{fig:pipeline}) designed to bridge the gap between task complexity and model competence so as to fully utilize these automatic tools.

\textbf{Stage I: We develop a suite of modality-specific expert models for healthy tissue inpainting}, built upon the FLUX.1-Fill-dev model. 
This strategy is based on the observation that generating healthy anatomical structures is more stable and predictable than generating lesions, as the former exhibits more tractable patterns and textures. For each modality, we curate a training dataset consisting exclusively of non-pathological samples from existing medical image repositories. Through parameter-efficient finetuning, these models learn to inpaint masked areas with high clinical accuracy. We further apply background restoration and edge blending to correct any unintended modifications made by the FLUX model outside the mask, ensuring the edited region blends seamlessly into the original image.

\textbf{Stage II: We leverage these expert models to modify lesion-bearing images ($L$) into their counterfactual `healthy' results ($H$).} 
Specifically, we fill the lesion area in $L$ using white pixels based on its ground-truth segmentation label. This masked image and its corresponding binary mask are then processed by the modality-matched expert model to synthesize $H$, where healthy tissue replaces lesion. 
Compared to distilling general-purpose generative models, our modality-specific approach not only restricts the high-variance generative process to a localized region to guarantee background consistency during the edit, but also ensures that tasks remain within the model’s learned distribution, thereby significantly reducing hallucinations. Furthermore, unlike manual data collection from the internet, our approach provides superior scalability and efficiency.

\textbf{Stage III: We implement a rejection sampling mechanism for the generated `healthy' images ($H$)} to further enhance the data quality within the Modification tasks. For modalities that resemble natural images (\textit{e.g.}, endoscopy and dermoscopy), we prompt the Qwen3-VL-32B-Instruct model \citep{qwen3vl} to filter out $H$ that still contain lesions, exhibit artifacts, or are of low quality. For other modalities, we train separate nnUNet models \citep{nnunet} for lesion segmentation and discard $H$ where lesions remain detectable. For further details, please refer to App.~\ref{app:auto_detail}.

\textbf{Stage IV: Triplet combination.} Using these high quality `lesion-healthy' counterfactual pairs, we generate diverse Modification task data by swapping $L$ and $H$ from niche of input and output and varying the textual prompts $P$. 

\subsubsection{Transformation}

Transformation tasks include a wide array of low-level medical image processing operations. 
Unlike the localized edits found in Perception and Modification categories, tasks in this category typically require a holistic transformation of the entire input image.

The rule-based construction pipeline of Transformation tasks is shown in Fig.~\ref{fig:pipeline}. From public repositories, we compile medical image pairs ($I$ and $O$) representing 11 distinct transformation targets under four typical low-level vision categories. We then design specialized textual prompts $P$ for each task to unify diverse medical image processing functions into the image editing framework.

\subsubsection{Post Processing}
\label{sec:post_process}

\textbf{Prompt rephrasing.} To enhance linguistic diversity, we utilize the Qwen-Max model to rephrase the prompts $P$ for each data triplet. 
We also illustrate the linguistic diversity of our prompts via a word cloud in Fig.~\ref{fig:dist}(b).

\textbf{Benchmark curation.} The training and test split of source dataset are strictly followed during the construction of \datasetname\ to preclude any data leakage.
Furthermore, we recruit three people with clinical background to \textbf{manually evaluate and curate} \numberofbenchdata\ of the most representative samples characterized by high clinical fidelity from raw data test split (Please refer to App.~\ref{app:manual_detail} for details) to serve as the benchmark of \datasetname, and we keep their original image size to minimize information loss. 

\textbf{Train split construction.} For train split, we establish three resolution bins (128, 256, and 512) and resize images to their nearest corresponding value.
To check the fidelity of training split, we randomly select 6,000 triplets for clinician evaluation, and \textbf{over 95\% are viewed as high quality}.

\subsection{\datasetname\ Evaluation}
\label{sec:evaluation}

We evaluate \datasetname\ through two approaches: (1) verifiable metrics for the Perception and Transformation tasks, amenable to reward design in prevailing reinforcement learning \citep{grpo, flowgrpo}; and (2) more subjective evaluations for the Modification tasks, reflecting their greater complexity.

\subsubsection{Verifiable Evaluation}

\textbf{Localization Accuracy Metric.} We use the \textbf{DICE} Score for evaluating the spatial overlapping performance in Perception tasks.
Notably, reconstructing a binary mask from the colored regions of an edited image is mathematically feasible when the background image and overlay color are known, and we detail this process in App.~\ref{app:mask_restore}, where
the average DICE between original and reconstructed masks is 0.999 for grayscale images and 0.970 for RGB images. 
This procedure is applied to both model’s output $O_M$ and the ground truth images $O$ to derive the mask of model's perceptual region and the ground truth region for DICE calculation.

To differentiate between models that accurately identify specific medical targets and those that merely generate coarse-grained masks, we further propose \textbf{Perception Accuracy}: The result is considered successful only if the DICE score exceeds a threshold of $\tau=0.8$. This metric allows us to analyze whether a model possesses the specialized medical knowledge required for image understanding.

\textbf{Image Similarity Metrics.} We utilize \textbf{PSNR} and \textbf{SSIM} \citep{ssim} to evaluate the similarity between the ground-truth and edited images at both the pixel and structural levels. For evaluations within the Perception perspective, we mask out the pixels corresponding to the ground-truth segmentation in both images. This allows us to specifically assess the model’s ability to preserve the background while performing the requested edit. Moreover, to test the clinical utility of models in transformation perspective, we also exemplify a task-based evaluation in App.~\ref{app:downstream_mar}.

\subsubsection{Evaluation for Modification Tasks}

\textbf{Vision-Language Model Rubric Scoring.}
Automating reliable assessments in the Modification tasks is inherently challenging, as edits are defined semantically and cannot be evaluated via deterministic rules. Existing benchmarks often leverage Vision-Language Models (VLMs) for this purpose, and we standardize the process and mitigate potential critic hallucinations by implementing a rubric-based scoring system. Specifically, we provide the VLM with the input image $I$, edit instruction $P$, reference output $O$, and the model’s generated result $O_{M}$. Guided by the rubric, the VLM then performs a holistic evaluation of $O_M$. 

We design a comprehensive scoring rubric (App. \ref{app:rubric}) that assesses both the fulfillment of the editing intent and the model's ability to preserve the background. We utilize GPT-5.2 as an automated evaluator for this process, and map the final score to $[0, 100]$. 

\textbf{Human Preference Ranking.} 
For each test case, we present the original triplet $(I, P, O)$ and the outputs of all tested models simultaneously to evaluators, who are then asked to rank the various model-generated results according to their preference.
By forcing this comparative ordering of all models, we are able to move beyond absolute quality scores and capture the relative strengths and weaknesses of current generative frameworks in a clinical setting.
Specifically, we recruit 3 evaluators with clinical backgrounds to assess and rank the images edited by the benchmarked models, and compute the average ranking. 

\textbf{Evaluation Agreement.} 
The Pairwise Spearman correlations among the three evaluators are $0.963$, $0.989$, and $0.957$ (all $p < 1\times10^{-5}$), demonstrating good inter-rater reliability. 
Moreover, the correlation between human preference ranking and VLM rubric score is $\rho = -0.905\ (p = 3\times 10^{-4})$. Since smaller Pref-Rank means better, the strong negative correlation confirms that the VLM evaluator aligns well with human clinical judgment.

\section{Experiments}
\label{sec:experiment}

\begin{table*}[t!]
\centering
\caption{\textbf{Overall result on \datasetname\ benchmark.} 
P-ACC means Perception Accuracy; B-PSNR and B-SSIM mean only calculate PSNR and SSIM on background pixels respectively; Rubric-S stands for the Rubric Score from VLM and Pref-Rank stands for human preference ranking.
Best values are marked in \textcolor{red}{red} while second bests are in \textcolor{blue}{blue}.}
\label{tab:overall}
\resizebox{\linewidth}{!}{
\begin{tabular}{l  c | c c c c | c c | c c}
\toprule
 & & \multicolumn{4}{c}{\textbf{Perception}} & \multicolumn{2}{|c|}{\textbf{Modification}} & \multicolumn{2}{c}{\textbf{Transformation}} \\ 
 \midrule
 \textbf{Model Name} & \textbf{Size} & \textbf{DICE} & \textbf{P-ACC} & \textbf{B-PSNR} & \textbf{B-SSIM} & \textbf{Rubric-S} & \textbf{Pref-Rank} & \textbf{PSNR} & \textbf{SSIM} \\ 
 \hline
\multicolumn{10}{c}{Open-Source} \\ \hline
SDXL-turbo \citep{sdxl_turbo} & 3.5B & 0.002 & 0.000 & \textcolor{blue}{16.6} & 0.467 & 8.4 & 7.7 & 15.4 & 0.369 \\
Bagel \citep{bagel} & 7B & 0.263 & 0.069 & 13.9 & 0.620 & 34.4 & 6.2 & 12.8 & 0.435 \\
OmniGen2 \citep{omnigen2} & 7B & 0.248 & 0.065 & 11.9 & 0.541 & 29.1 & 7.1 & 9.1 & 0.305 \\
Step1X-Edit \citep{step1xedit} & 21B & 0.332 & 0.126 & 15.5 & \textcolor{blue}{0.727} & 35.6 & 4.5 & 16.0 & 0.533 \\
Qwen-Image-Edit \citep{qwenimage} & 27B & 0.387 & 0.153 & 15.4 & 0.722 & 32.2 & 5.5 & 18.8 & 0.598 \\
FLUX.1-Kontext-dev \citep{flux1kontext} & 12B & 0.341 & 0.126 & 15.4 & 0.701 & 37.8 & 6.2 & 17.3 & 0.514 \\
\textbf{OmniGen2-MIE (Ours)} & 7B & \textcolor{red}{\textbf{0.831}} & \textcolor{red}{\textbf{0.737}} & \textcolor{red}{\textbf{28.1}} & \textcolor{red}{\textbf{0.917}} & \textcolor{red}{\textbf{65.9}} & \textcolor{red}{\textbf{1.4}} & \textcolor{red}{\textbf{22.2}} & \textcolor{red}{\textbf{0.668}} \\ \hline
\multicolumn{10}{c}{Closed-Source} \\ \hline
GPT-Image-1 \citep{gptimage1} & & \textcolor{blue}{0.467} & \textcolor{blue}{0.221} & 16.3 & 0.510 & 42.8 & 4.8 & 14.2 & 0.430 \\
Nano Banana Pro \citep{gemini3pro} & & 0.426 & 0.202 & 12.8 & 0.413 & \textcolor{blue}{63.4} & \textcolor{blue}{2.0} & \textcolor{blue}{19.8} & \textcolor{blue}{0.639} \\
Imagen4 \citep{imagen4} & & 0.142 & 0.000 & 8.9 & 0.210 & 19.7 & 7.4 & 8.0 & 0.171 \\ \bottomrule
\end{tabular}
}
\end{table*}

\subsection{Baselines}

We evaluate nine models on \datasetname, comprising six open-source models: Qwen-Image-Edit-2511 \citep{qwenimage}, Bagel \citep{bagel}, OmniGen2 \citep{omnigen2}, Step1X-Edit-v1p2 \citep{step1xedit}, FLUX.1-Kontext-dev \citep{flux1kontext} and SDXL-turbo \citep{sdxl_turbo}, plus three closed-source models: Nano Banana Pro \cite{gemini3pro}, GPT-Image-1 \citep{gptimage1}, and Imagen4 \citep{imagen4}. We implement open-source models following their official inference settings.

To validate the effectiveness of \datasetname, we finetune the OmniGen2 baseline on the training split and subject it to the same evaluation protocol as the other models. Specifically, we train the Diffusion Transformer (DiT) component for 20,000 iterations, employing a global batch size of 64 and a learning rate of 1e-4. We also finetune FLUX.1-Kontext baseline model to demonstrate the generality of our dataset across different model architectures. Please refer to App.~\ref{app:flux_finetune} for details. 

\subsection{Quantitative Results}

\begin{figure*}[t!]
    \centering
    \includegraphics[width=0.9\linewidth]{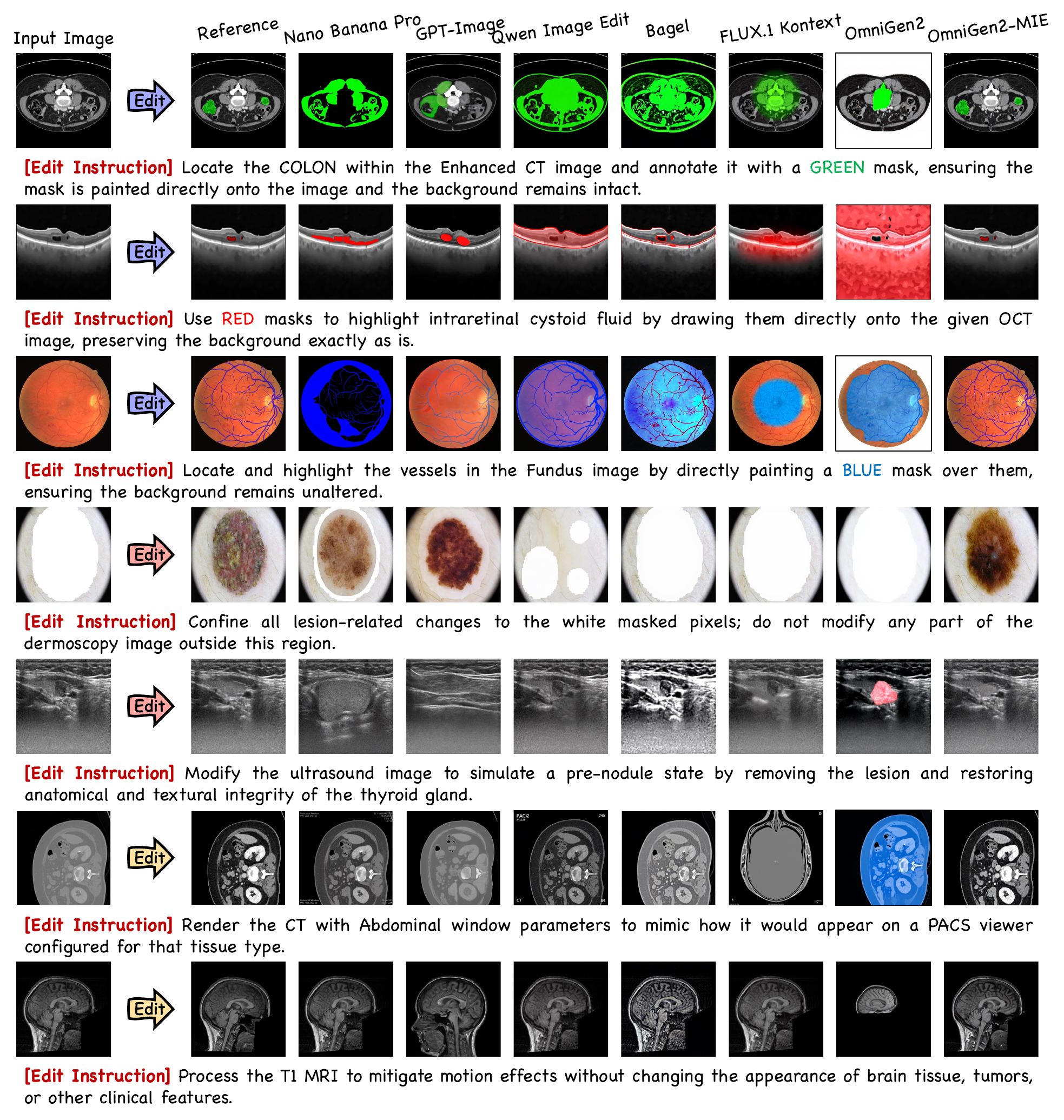}
    \caption{\textbf{Qualitative editing result comparison.}}
    \label{fig:qualitative}
\end{figure*}

We report the benchmarking results of \datasetname\ in Tab.~\ref{tab:overall}. First, the extremely low \textbf{perception accuracy} indicates that \textbf{all tested models except ours} fail to accurately comprehend and localize the specified anatomical targets under our evaluation protocol. Consequently, in \textbf{Modification tasks}, most of them are unable to generate clinically meaningful edits. Although a few models, such as Nano Banana Pro, achieve competitive results, we are indeed observing the 'right-for-the-wrong-reason' phenomenon, a risk that must be strictly avoided in clinical settings. Since the poor performance in Perception tasks exposes their intrinsic lack of necessary medical knowledge, their edits cannot be justified.
Notably in \textbf{Transformation tasks}, Nano Banana Pro also presents competitive results in certain cases. This may be attributed to the similarity between tasks like denoising or artifact removal and general-purpose low-level vision tasks, for which the model already possesses some capability \citep{nanobananaprolowlevel}. Alternatively, it is possible that similar medical image processing tasks were included in its training set. Regardless, its absolute performance remains insufficient for practical clinical deployment.
\textbf{In summary, the benchmark result demonstrates that current multimodal generative model cannot meet the requirement of medical imaging editing.}

\textbf{Conversely, after training on \datasetname, a standard baseline model can achieve superior medical editing capabilities.} As shown in Tab.~\ref{tab:overall}, the OmniGen2-MIE model delivers the best performance across all three editing perspectives. The most significant improvements are observed in the Perception perspective, which demonstrate that \datasetname\ can effectively inject essential medical knowledge, thereby enhancing the interpretability of downstream editing tasks. Furthermore, in the Modification and Transformation tasks, where general-purpose editing abilities transfer more readily, our enhanced model still yields superior editing results compared to Nano Banana Pro.
These findings highlight the pivotal role of our dataset in domain adaptation and establish a foundation for the development of understanding-generation unified medical models.

\subsection{Qualitative Results}

Fig.~\ref{fig:qualitative} presents qualitative editing results for several baseline models across the diverse modalities and tasks in \datasetname. These results demonstrate that the finetuned model exhibits an enhanced capability in both understanding and generation, allowing it to navigate the inherent complexities of medical image editing. 
Moreover, despite being explicitly prompted, even sophisticated closed-source models such as Nano Banana Pro fail to maintain background consistency in certain tasks. While their instruction-following proficiency stems from large-scale pre-training on natural image pairs, these capabilities tend to degrade when the distribution of medical modalities deviates significantly from the natural images seen during pre-training. To further study the impact of modality deviation, we conduct a modality-wise analysis in App.~\ref{app:mod_dev}, and the results prove our judgment. This observation underscores the necessity of a highly diverse dataset like \datasetname\ to equip models with the capacity to handle a vast range of medical imaging modalities.

\begin{wraptable}{r}{0.5\linewidth}
\centering
\vspace{-0.5cm}
\caption{\textbf{Ablation study result on \datasetname.} \textbf{P} stands for Perception, \textbf{M} stands for Modification, and \textbf{T} stands for Transformation. Best values are marked in \textcolor{red}{red}, second bests are in \textcolor{blue}{blue}.}
\label{tab:ablation}
\resizebox{\linewidth}{!}{
\begin{tabular}{l | c c |  c | c c}
\toprule
 & \multicolumn{2}{c}{\textbf{Perception}} & \multicolumn{1}{c}{\textbf{Modification}} & \multicolumn{2}{c}{\textbf{Transformation}} \\ 
\midrule
\textbf{Training Data} & \textbf{DICE} & \textbf{ACC} & \textbf{RubricScore} & \textbf{PSNR} & \textbf{SSIM} \\ 
\midrule
\textbf{Baseline} (No train) & 0.248 & 0.065 & 29.1 & 8.3 & 0.280 \\
\textbf{P}-only & \textcolor{red}{\textbf{0.833}} & \textcolor{red}{\textbf{0.740}} & 37.8 & 19.2 & 0.609 \\
\textbf{M}-only & 0.001 & 0.000 & \textcolor{blue}{57.5} & 19.1 & 0.606 \\
\textbf{T}-only & 0.034 & 0.000 & 15.0 & \textcolor{red}{\textbf{23.9}} & \textcolor{red}{\textbf{0.696}} \\
\midrule
\textbf{\datasetname} & \textcolor{blue}{0.831} & \textcolor{blue}{0.737} & \textcolor{red}{\textbf{65.9}} & \textcolor{blue}{22.2} & \textcolor{blue}{0.668} \\ \bottomrule
\end{tabular}
}
\end{wraptable}

\subsection{Ablation Study}

To investigate the contribution of each task category, we conduct an ablation study by training models on individual perspective of \datasetname. 
We again utilize OmniGen2 as baseline model, following the training recipe described above while varying only the training data. 
As shown in Tab~\ref{tab:ablation}, each specialized model significantly outperforms the original baseline in its respective domain, validating the high information density and clinical relevance of our data. 

\begin{wrapfigure}{r}{0.43\textwidth}
    \vspace{-1cm}
    \centering
    \includegraphics[width=\linewidth]{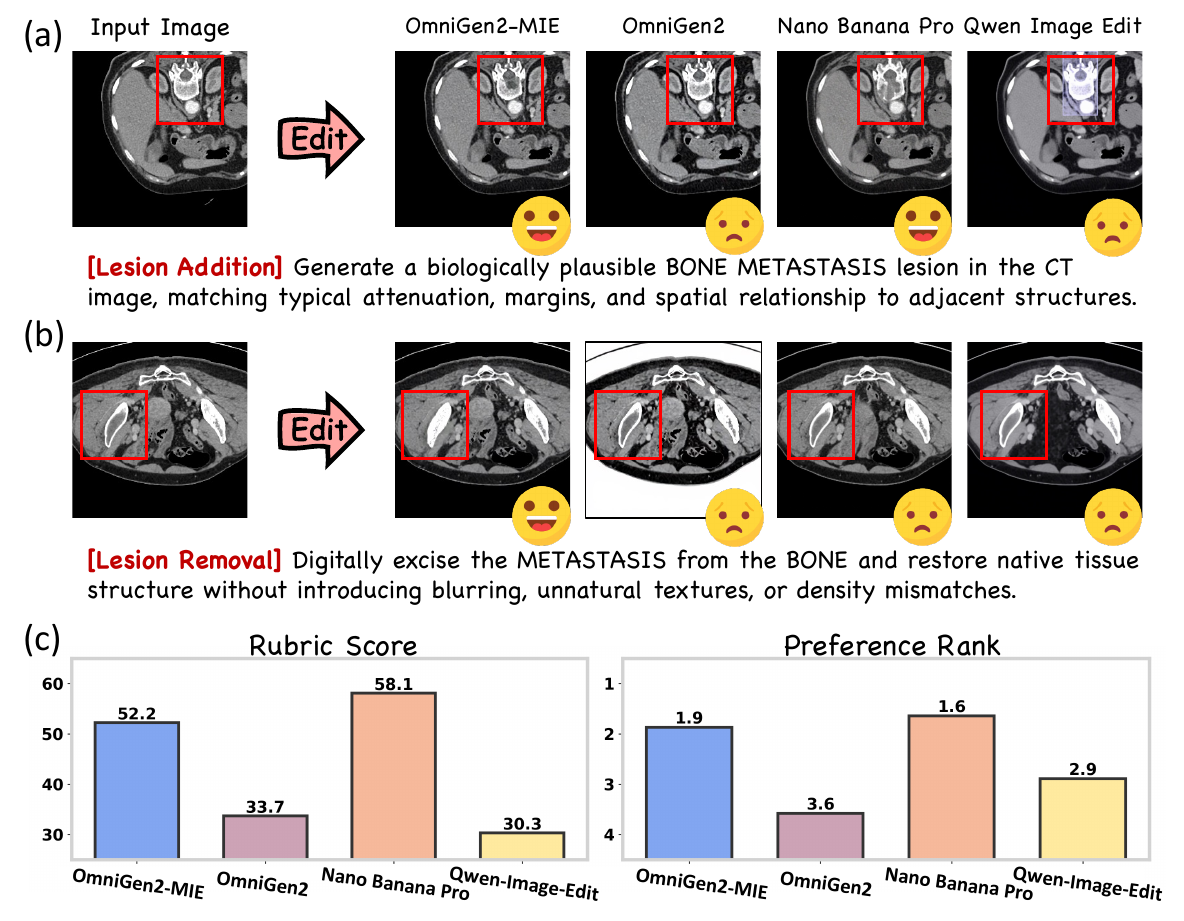}
    \caption{\textbf{Generalization test assessment.} (a) and (b): Edit samples output by different models on bone metastasis addition (a) and removal (b) tasks. Red bounding boxes are  added post-hoc to highlight the edited regions for visualization; (c): Quantitative assessments following the recipe of Modification task evaluation.}
    \label{fig:generalization}
    \vspace{-1cm}
\end{wrapfigure}

For the model trained on the full dataset, it achieves comparable or even better performance on all three perspectives, showing the effectiveness of the joint training.
More importantly, we observe significant performance improvement in the Modification perspective, demonstrating visual understanding ability has the potential to enhance visual generation ability.
In summary, the ablation study shows that \textbf{\datasetname\ can provide a synergistic training signal, enabling the development of a versatile model capable of handling diverse medical editing tasks simultaneously.}

\subsection{Generalization Test}
\label{sec:generalization}

To further investigate the cross-task synergy and the resulting generalization capabilities, we conduct an out-of-distribution (OOD) editing experiment. Specifically, we target `bone metastasis', a medical target included in Perception tasks but strictly excluded from the Modification training data. We then prompt the OmniGen2-MIE model to perform metastasis addition and removal in CT scans. 


As shown in Fig.~\ref{fig:generalization}, OmniGen2-MIE significantly outperforms OmniGen2 on this unseen task, demonstrating that our unified training on \datasetname\ can enhance the model's generalization capabilities across editing tasks.
We also observe that Nano Banana Pro achieves the best OOD editing performance, marginally surpassing OmniGen2-MIE.
We attribute this performance to the utilization of massive-scale general and medical editing data, which further underscores the necessity of scaling up medical editing data.

\section{Conclusion and Limitations}

In this paper, we introduce \datasetname, a large-scale and diverse dataset for text-guided medical image editing. By unifying Perception, Modification, and Transformation tasks into the paradigm of editing, our dataset bridges the gap between medical image understanding and generation. We develop a robust curation pipeline, integrating modality-specific expert models with rule-based synthesis, and enforce rigorous manual quality control to ensure clinical fidelity across all data. Extensive benchmarking demonstrates that model trained on \datasetname\ consistently outperforms both SOTA open-source and proprietary multimodal models while exhibiting exceptional generalization to unseen clinical tasks. 
Our work thus provides the data foundation to support the development and evaluation of multimodal generative models for clinical applications.

Limitations of our work include the inability to capture all medical imaging modalities, the scarcity of rare clinical cases, and the current focus on editing tasks alone.

\ifdefined\nipsanonymous

\else
\section*{Acknowledgment}
This work was supported by Damo Academy through Damo Academy Research Intern Program.
\fi

\bibliographystyle{plain}
\bibliography{miedb}

\newpage
\appendix
\include{App/appendix}


\ifdefined\nipsanonymous
\newpage
\input{checklist.tex}
\else

\fi

\end{document}

%% file: App/appendix.tex
\etocdepthtag.toc{appendix}
\section*{Appendix Table of Contents}
\vspace{0.5em}
{
\etocsettagdepth{mainpart}{none}
\etocsettagdepth{appendix}{subsubsection}
\tableofcontents
}

\newpage
\section{Data Sources}
\label{app:data_list}

Our work is compiled based on following public medical image repositories: 

\input{App/data_lisences}

We also appreciate MedSegBench\citep{medsegbench} and MedSegDB\citep{medsegdb} for collecting and pre-processing some of these datasets.

\newpage
\section{Construction Details}
\label{app:construct_detail}

\subsection{Processing Flowchart}

\begin{figure}[h!]
    \centering
    \includegraphics[width=\linewidth]{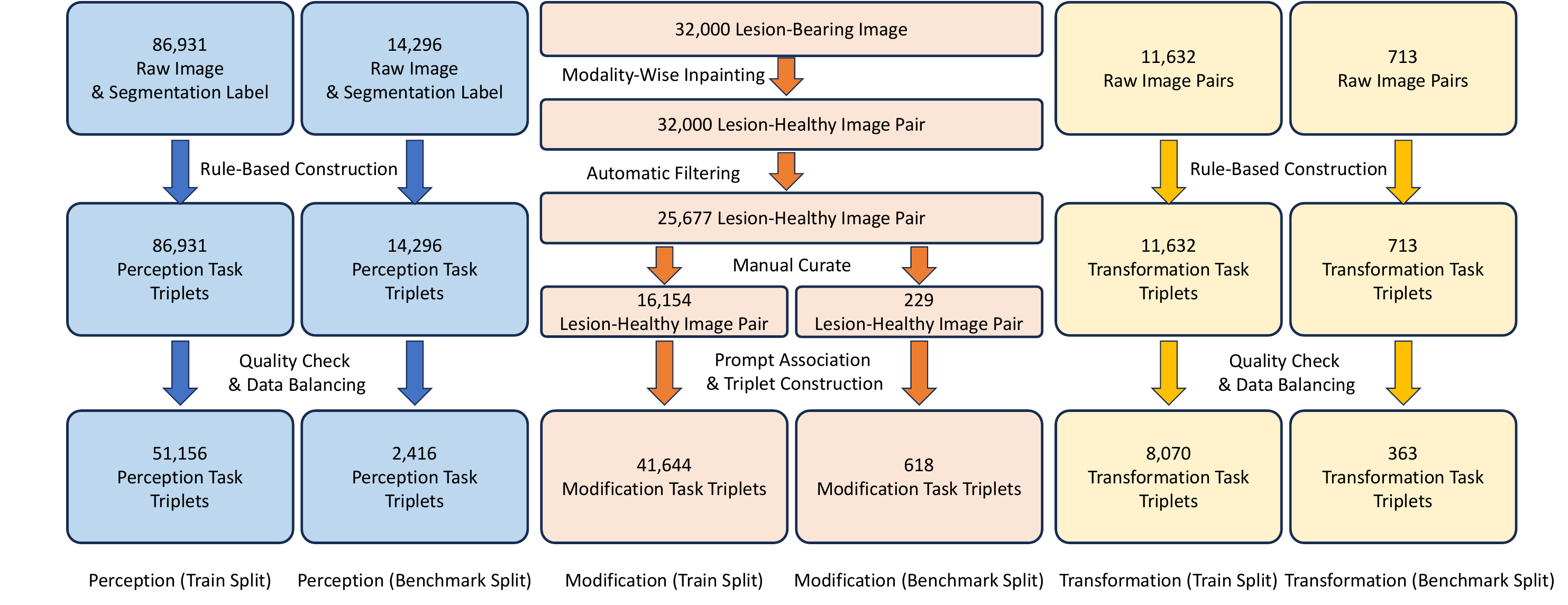}
    \caption{\textbf{Construction details of three perspective.} We manually curate the benchmark split to uphold high clinical standards. The remaining training data is validated through sampling-based quality checks, establishing a high-quality data proportion exceeding 95\%.}
\end{figure}

\subsection{Inpainting Model Training Details}

We use LoRA finetuning to train FLUX.1-Fill-dev. Implementation details of each modality experts are listed in Tab.~\ref{tab:expert_detail}. We conduct a simple hyper-parameter search over learning rate and number of training steps, selecting the final model based on both empirical inpainting performance and automatic rejection rate.

\begin{table}[h!]
\centering
\small
\resizebox{\linewidth}{!}{
\begin{tabular}{l|llllll}
\toprule
\textbf{Hyper-Parameters} & \textbf{CT} & \textbf{Ultrasound} & \textbf{MRI} & \textbf{Xray} & \textbf{Dermoscopy} & \textbf{Endoscopy}\\ \midrule
Finetuning method & LoRA & LoRA & LoRA & LoRA & LoRA & LoRA\\
Rank & 64 & 64 & 64 & 64 & 64 & 64 \\
Steps & 5,000 & 2,000 & 5,000 & 2,000 & 5,000 & 2,000 \\
\#GPUs & 4 & 4 & 4 & 4 & 4 & 4\\
Per-device batch size & 4 & 4 & 4 & 4 & 4 & 4 \\
Global batch size (effective) & 16 & 16 & 16 & 16 & 16 & 16 \\
Learning rate & $1 \times 10^{-4}$ & $1 \times 10^{-4}$ 
& $1 \times 10^{-5}$ & $1 \times 10^{-5}$
& $1 \times 10^{-5}$ & $1 \times 10^{-4}$\\
Precision & BF16 & BF16 & BF16 & BF16 & BF16 & BF16\\ \bottomrule
\end{tabular}
}
\caption{\textbf{Training hyper-parameters used for finetuning FLUX.1-Fill-dev into modality specific health-tissue inpainting expert.}}
\label{tab:expert_detail}
\end{table}

\subsection{Automatic Filtering Details}
\label{app:auto_detail}

\subsubsection{nnUNet Filtering}

We use the default standard 5-fold cross-validation for nnUNet training following its official implementations. 
The model structure and training hyper-parameters are generated automatically via \verb|nnUNetv2_plan_and_preprocess| tool. To prevent possible data leakage, we preclude the nnUNet 5-fold cross-validation training data from lesion-bearing images to be inpainted in Modification Stage II. 
The model is trained to generate segmentation map for given medical images. During rejection sampling, the sample is discarded if nnUNet still detects lesion-like pixels in the inpainted region.

\subsubsection{Qwen3-VL-32B-Instruct Filtering}

For each generated sample, we compose a three-panel collage consisting of: (1) the \textbf{inpainting mask} (left), which highlights the edited region in white; (2) the \textbf{edited image} $H$ (center), where the lesion has been replaced with synthetic healthy tissue; and (3) the \textbf{original image} $L$ (right), serving as a reference. 
To further aid the model's assessment, the edited region in the center frame is also highlighted with a red bounding box.
Then, the collaged image, together with a modality-specific structured prompt, is fed to Qwen3-VL-32B-Instruct. The prompt instructs the model to evaluate the edited region on multiple clinically relevant dimensions and return a structured JSON response.

Based on the model output, we retain only the high-quality samples. Specifically, a generated image $H$ is \textbf{accepted} only if its image quality is rated as ``good'' and no residual lesion or unnatural content is detected (rated as ``no''). All other samples are \textbf{discarded}. This conservative filtering strategy ensures that only clinically plausible inpainting results enter the subsequent triplet combination stage.

Below we showcase the VLM prompt for Endoscopy filtering:

\input{App/vlm_filter}

\subsection{Manual Inspection Details}
\label{app:manual_detail}

We engage \textbf{three board-certified medical experts} to assess the images in MieDB-100k. The evaluation protocol differs by task type, reflecting the distinct nature of each data source:

\textbf{Perception and Transformation tasks (rule-based data).} We sample approximately 3,000 cases from the source test split and removed those with mask mismatches or transformation errors. Owing to the high quality of the source datasets, only a negligible number of samples are excluded. The remaining data constitute the MieDB-100k benchmark split for these tasks.

\textbf{Modification tasks (model-generated data).} All samples that passed the rejection-sampling stage are submitted to expert review to ensure clinical validity. The experts are informed of the intended use of the data (i.e., serving as medical image editing pairs) and are asked to classify each sample into three categories:
  \textbf{Good}. High-quality edits that successfully accomplish the intended modification without noticeable artifacts;
  \textbf{Fair}. Edits that achieve the intended modification with only slight, clinically negligible traces that do not affect their usability as medical editing data;
  \textbf{Poor}. Failed edits or samples with obvious artifacts or incorrect modifications.

Note that to support consistent evaluation, we present each case as a collage of the original image and the inpainted image, and provide a detailed instruction document including task-specific descriptions and representative examples.

\textbf{Quality-check results.} 
Final labels are determined by \textbf{majority vote} among the three experts, and more than $76\%$ of the reviewed samples receive unanimous 3-out-of-3 agreement, indicating strong inter-rater consistency.
The results are approximately evenly distributed across the three categories. We remove all \textit{poor} samples and construct the benchmark split using only \textit{good} samples. In addition, we randomly sample 6,000 cases from the full 100k training set for spot-checking, and find that over $95\%$ are of high quality.

\subsection{Preservation Rates}

\begin{table}[h!]
\centering
\small
\resizebox{\linewidth}{!}{
\begin{tabular}{l|cccccc}
\toprule
\textbf{Phase} & \textbf{CT} & \textbf{Ultrasound} & \textbf{MRI} & \textbf{Xray} & \textbf{Dermoscopy} & \textbf{Endoscopy}\\ \midrule
Automatic Filtering & 86.1\% & 76.1\% & 61.6\% & 88.2\% & 64.9\% & 67.6\% \\
Manual Inspection & 64.0\%	&64.5\%	&49.9\%	&55.3\%	&73.2\%	&66.0\%\\
\bottomrule
\end{tabular}
}
\caption{\textbf{Data preservation rates during automatic filtering and manual inspection.}}
\end{table}

\newpage
\subsection{Target Distribution}

\begin{figure}[h!]
    \centering
    \includegraphics[width=0.85\linewidth]{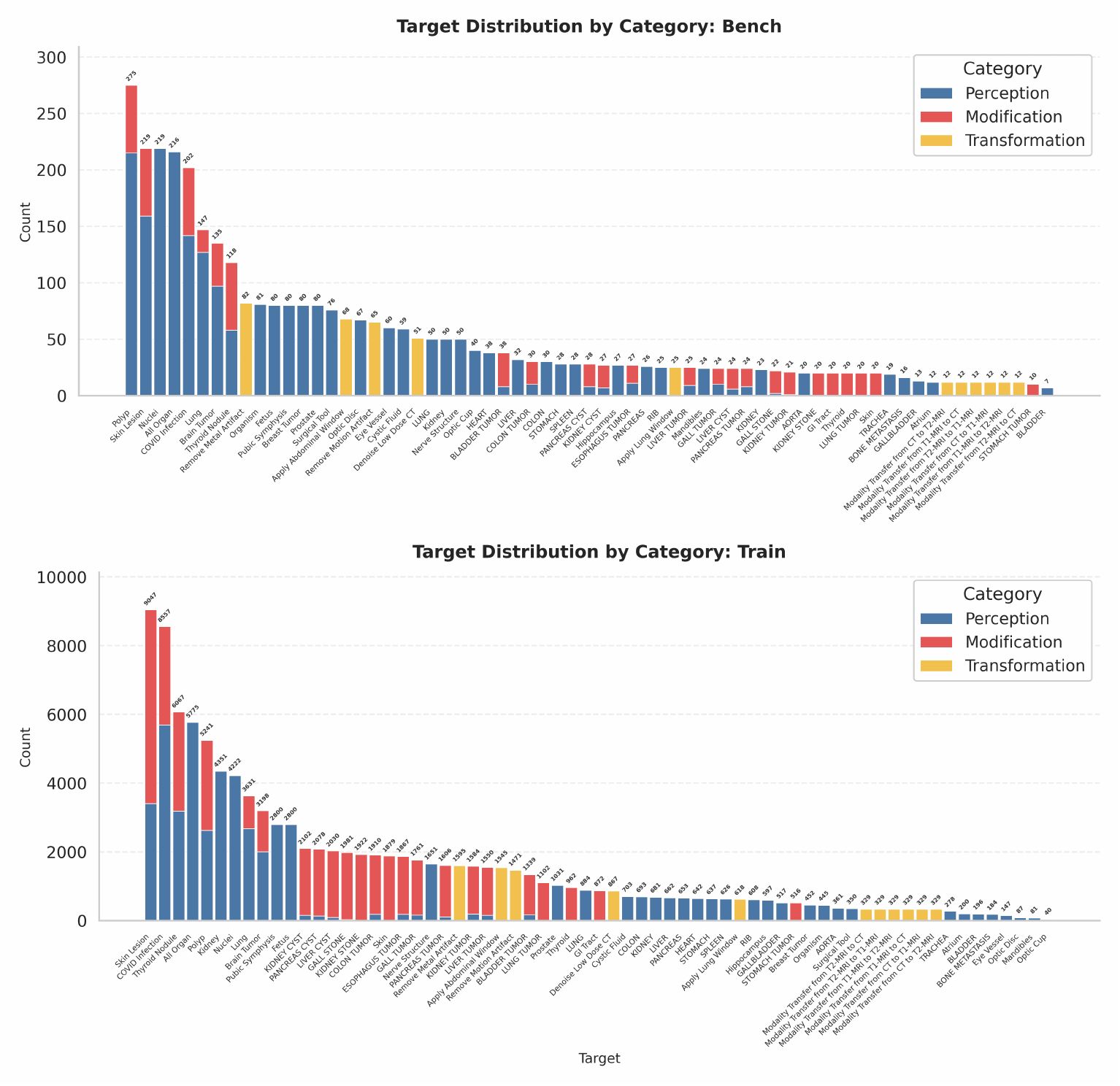}
    \caption{\textbf{Distribution details of edit target.}}
\end{figure}

\section{Implementation Details of OmniGen2-MIE}
\label{app:train_detail}

\begin{table}[h!]
\centering
\small
\begin{tabular}{ll}
\toprule
\textbf{Hyper-Parameter} & \textbf{Value} \\ \midrule
Finetuning method & Full-Parameter Finetuning \\
snr\_type & lognorm \\
do\_shift & True \\
dynamic\_time\_shift & True \\
Steps & 20, 000 \\
\#GPUs & 8 $\times$ H20 (96G) \\
Per-device batch size & 8 \\
Gradient accumulation & 1 \\
Global batch size (effective) & 64 \\
Learning rate & $1 \times 10^{-4}$ \\
LR scheduler & timm\_constant\_with\_warmup \\
Warm-up\_t & 500 \\
Precision & BF16 \\
Random seed & 2233 \\ \bottomrule
\end{tabular}
\caption{\textbf{Training hyper-parameters used for finetuning OmniGen2-MIE on our dataset.}}
\label{tab:training-params}
\end{table}

\clearpage
\newpage
\section{Evaluation Details}
\label{app:eval_detail}

\subsection{Mask Reconstruction via Alpha De-blending}
\label{app:mask_restore}

\subsubsection{Mathematics}
To recover the segmentation mask from the visualized output, we model the edited image $\mathbf{O}$ as a linear interpolation between the original background image $\mathbf{B}$ (a.k.a. the input image $I$) and a known overlay color $\mathbf{C}$ (red, green or blue). This relationship is governed by the per-pixel alpha channel $\alpha \in [0,1]$, according to the standard alpha blending equation:
\begin{equation}
\mathbf{O} = (1 - \alpha)\mathbf{B} + \alpha\mathbf{C}
\end{equation}
By rearranging the terms as $\mathbf{O}-\mathbf{B}=\alpha(\mathbf{C}-\mathbf{B})$, the scalar value $\alpha$ can be interpreted as the projection of the observed color shift onto the vector representing the maximum possible color change. To account for potential noise in the RGB space, we solve for $\alpha$ at each pixel using the least-squares solution:
\begin{equation}
\alpha = \frac{(\mathbf{O} - \mathbf{B}) \cdot (\mathbf{C} - \mathbf{B})}{|\mathbf{C} - \mathbf{B}|^2}
\end{equation}
The continuous alpha map is subsequently binarized to produce the final segmentation mask $M$. This is achieved by applying a global threshold $\tau$, such that:
\begin{equation}
M_{i,j} =
\begin{cases}
1 & \text{if } \alpha_{i,j} > \tau \\
0 & \text{otherwise}
\end{cases}
\end{equation}
In our implementation, a threshold of $\tau=0.5$ is utilized to effectively separate the predicted regions from the background.
The average DICE between original and reconstructed masks is 0.999 for grayscale images (e.g., X-ray) and 0.970 for RGB images (e.g., Endoscopy), confirming robustness to compression and color drift.

\subsubsection{Case of mask reconstruction}

\begin{figure}[!h]
    \centering
    \includegraphics[width=0.5\linewidth]{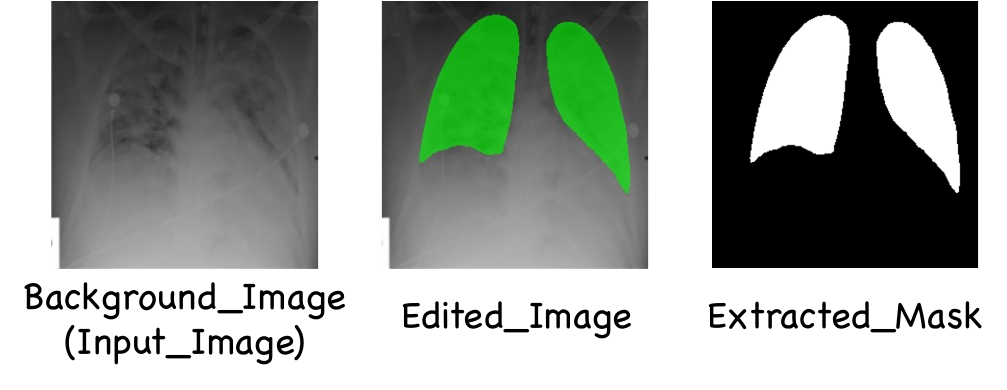}
    \caption{\textbf{Case of perception mask reconstruction.}}
\end{figure}

\subsection{VLM Automatic Scoring}
\label{app:rubric}

\subsubsection{VLM Scoring Rubric}
\input{App/rubric}

\subsubsection{Case}

\begin{figure}[h]
    \centering
    \includegraphics[width=\linewidth]{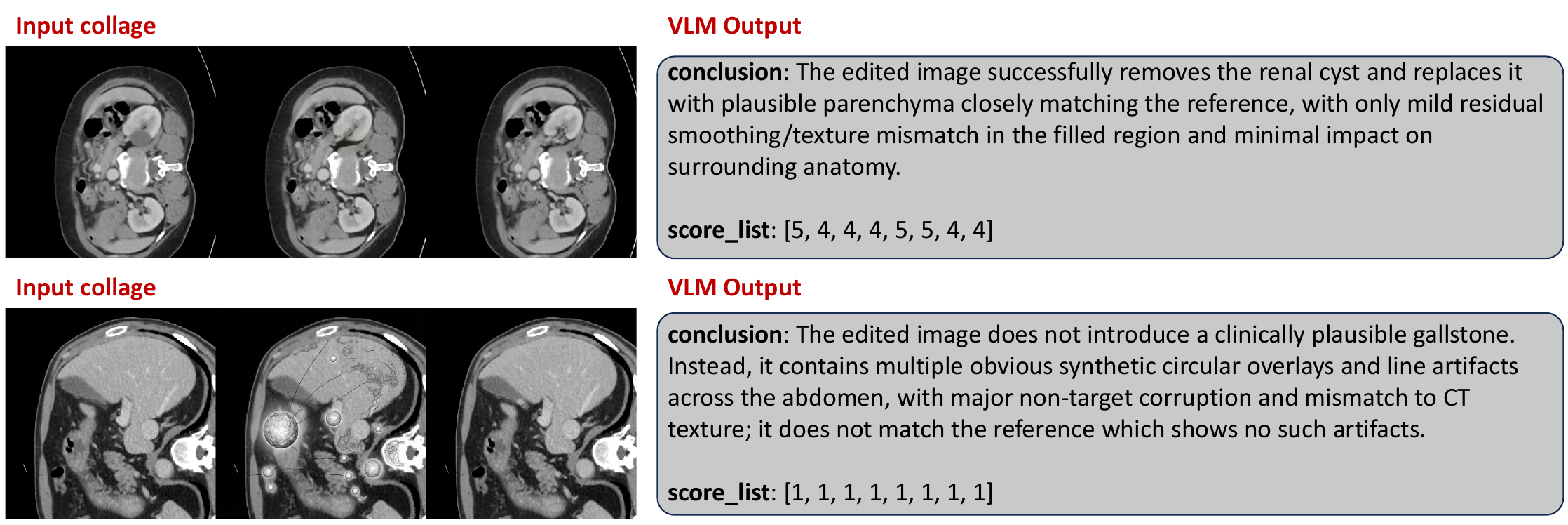}
    \caption{\textbf{Cases of VLM rubric scoring.}}
\end{figure}

\subsubsection{Alternative VLM Consistency}
\label{app:cross_evaluator}

To validate the stability and robustness of our VLM rubric scoring methodology, we evaluate the Modification task outputs using two different VLM evaluators: Gemini-3-flash and GPT-5.2. As shown in Tab.~\ref{tab:cross_evaluator}, despite the absolute scores varying between evaluators, the relative ranking of all benchmarked models remains highly consistent. The Spearman rank correlation between the two evaluators is $\rho = 0.951\ (p = 2\times10^{-5})$, indicating strong agreement. This cross-evaluator consistency demonstrates that our rubric-based evaluation protocol is robust to the choice of VLM backbone and reliably reflects the relative quality of different models' editing outputs.

\begin{table}[h]
\centering
\caption{\textbf{VLM Rubric Score using different evaluators.}}
\label{tab:cross_evaluator}
\resizebox{\linewidth}{!}{
\begin{tabular}{l | c c c c c c c c c c}
\toprule
\textbf{Evaluator} & \textbf{OmniGen2-MIE} & \textbf{Gemini} & \textbf{GPT-Image} & \textbf{FLUX} & \textbf{Step1x} & \textbf{Bagel} & \textbf{Qwen-Image} & \textbf{OmniGen2} & \textbf{Imagen4} & \textbf{SDXL} \\
\midrule
Gemini-3-flash & 37.0 & 35.9 & 26.1 & 18.6 & 21.0 & 19.8 & 10.8 & 13.4 & 8.9 & 7.3 \\
GPT-5.2 & 65.9 & 63.4 & 42.8 & 37.8 & 35.6 & 34.4 & 32.2 & 29.1 & 19.7 & 8.4 \\
\bottomrule
\end{tabular}
}
\end{table}

\clearpage
\newpage
\section{Supplementary Experiments}

\subsection{Modality-Wise Performance Analysis}
\label{app:mod_dev}

\begin{figure}[h]
    \centering
    \includegraphics[width=0.8\linewidth]{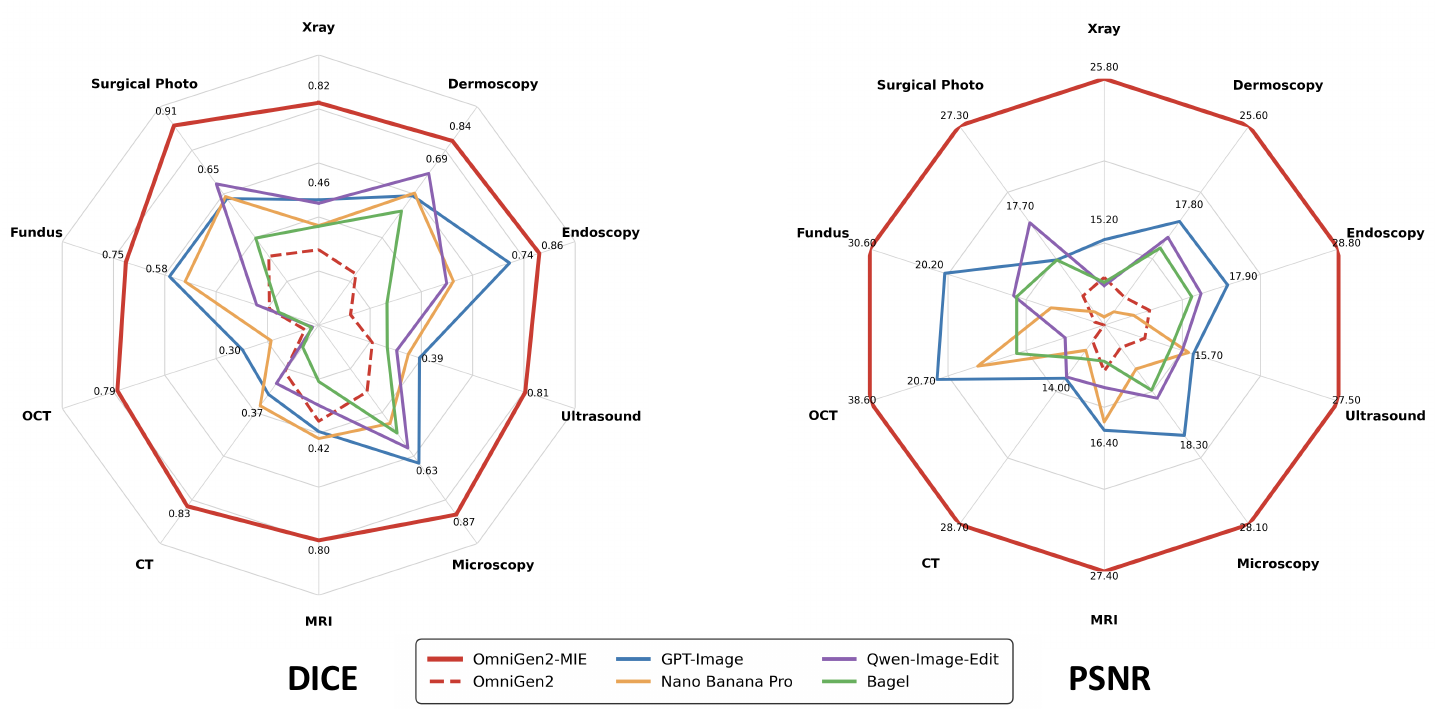}
    \caption{\textbf{Modality-wise performance analysis result within perception perspective.} Left: DICE score; right: PSNR score.}
    \label{fig:radar}
\end{figure}

To investigate the impact of modality deviation, we conduct a modality-wise analysis of the benchmarking results within the Perception perspective. Specifically, we report the DICE and PSNR scores of six representative models across all medical imaging modalities included in MieDB-100k. As illustrated in Fig.~\ref{fig:radar}, the experimental results are consistent with our hypotheses. For the baseline models, performance is unevenly distributed across the various modalities: They achieve relatively strong results on modalities that resemble natural images, such as Endoscopy, Dermoscopy, and Surgical Photo. However, on non-optical modalities (e.g., CT, MRI, Ultrasound), their performance degrades drastically. In contrast, the model trained on our dataset exhibits balanced and superior performance across all imaging types. Collectively, these results demonstrate that a diverse dataset like MieDB-100k is essential for successfully adapting multi-modal generative models to the medical domain.

\subsection{Multi-Round Generation}
\label{app:multi_round}

\begin{table}[h]
\centering
\caption{\textbf{Multi-round generation result.} Best values are marked in \textbf{Bold}}
\label{tab:pass3}
\resizebox{\linewidth}{!}{
\begin{tabular}{l | cc cc cc cc | cc cc}
\toprule
 & \multicolumn{8}{c}{\textbf{Perception}} & \multicolumn{4}{c}{\textbf{Transformation}} \\ 
\cline{2-13}
 & \multicolumn{2}{c}{DICE} & \multicolumn{2}{c}{P-ACC} & \multicolumn{2}{c}{B-PSNR} & \multicolumn{2}{c}{B-SSIM} & \multicolumn{2}{c}{PSNR} & \multicolumn{2}{c}{SSIM} \\
 & Pass@1 & Pass@3 & Pass@1 & Pass@3 & Pass@1 & Pass@3 & Pass@1 & Pass@3 & Pass@1 & Pass@3 & Pass@1 & Pass@3 \\
\hline \hline
\multicolumn{13}{c}{Open-Source} \\ \hline
SDXL-turbo & 0.002 & 0.003 & 0.000 & 0.000 & 16.6 & 17.0 & 0.467 & 0.484 & 15.4 & 15.7 & 0.369 & 0.396 \\
Bagel & 0.263 & 0.383 & 0.069 & 0.137 & 13.9 & 16.1 & 0.620 & 0.703 & 12.8 & 15.2 & 0.435 & 0.533 \\
OmniGen2 & 0.248 & 0.357 & 0.065 & 0.125 & 11.9 & 14.4 & 0.541 & 0.628 & 9.1 & 15.3 & 0.305 & 0.528 \\
Step1X-Edit & 0.332 & 0.369 & 0.126 & 0.143 & 15.5 & 16.4 & 0.727 & 0.748 & 16.0 & 16.5 & 0.533 & 0.548 \\
FLUX.1-Kontext-dev & 0.347 & 0.41 & 0.126 & 0.174 & 15.4 & 16.5 & 0.701 & 0.761 & 17.3 & 18.9 & 0.514 & 0.578 \\
Qwen-Image-Edit & 0.387 & 0.493 & 0.153 & 0.249 & 15.4 & 17.4 & 0.722 & 0.795 & 18.8 & 20.2 & 0.598 & 0.645 \\ \midrule
\textbf{OmniGen2-MIE (Ours)} & \textbf{0.831} & \textbf{0.856} & \textbf{0.737} & \textbf{0.789} & \textbf{28.1} & \textbf{28.8} & \textbf{0.917} & \textbf{0.921} & \textbf{22.2} & \textbf{23.0} & \textbf{0.668} & \textbf{0.697} \\
\bottomrule
\end{tabular}
}
\end{table}

To mitigate the inherent variance of the generative process, we report Pass@3 scores for the open-source models on Perception tasks. Specifically, we generate three independent outputs for each editing task and select the highest-performing sample to represent the task’s score. These results are then averaged across all tasks to provide a robust assessment of overall performance.

The results of the multi-round generation tests are summarized in Tab.~\ref{tab:pass3}. While multi-round generation improves the absolute scores for baseline models, it does not alter the underlying fact that these models lack essential medical knowledge. Furthermore, the significant fluctuations across rounds expose the high-variance nature of these baselines, undermining their reliability under clinical applications. In contrast, our model exhibits remarkable stability across all three trials. This consistency suggests that model trained on \datasetname\ has developed a deterministic understanding of medical concepts rather than relying on fortuitous generation.

\subsection{Baseline Model Generalization}
\label{app:flux_finetune}

Besides OmniGen2, we also finetune FLUX.1-Kontext\citep{flux1kontext} baseline model on \datasetname\ to demonstrate the generality of our dataset across different model architectures. With global batch of 64, we iterate the backbone transformer for 40,000 steps under learning rate of $5\times10^{-6}$ using AdamW optimizer. As shown in Tab.~\ref{tab:flux_finetune}, the finetuned FluxKontext-MIE model also achieves substantial improvements over the original FluxKontext baseline across all three editing perspectives. These results confirm that the benefits of training on \datasetname\ are not limited to a single architecture, but generalize across different multimodal generative models.

\begin{table}[h]
\centering
\caption{\textbf{Result of finetuning FLUX.1-Kontext on \datasetname.} Best values are marked in \textcolor{red}{red}.}
\label{tab:flux_finetune}
\resizebox{0.9\linewidth}{!}{
\begin{tabular}{l | c c c c | c | c c}
\toprule
 & \multicolumn{4}{c|}{\textbf{Perception}} & \multicolumn{1}{c|}{\textbf{Modification}} & \multicolumn{2}{c}{\textbf{Transformation}} \\ 
\midrule
\textbf{Model} & \textbf{DICE} & \textbf{P-ACC} & \textbf{B-PSNR} & \textbf{B-SSIM} & \textbf{Rubric-S} & \textbf{PSNR} & \textbf{SSIM} \\ 
\midrule
FluxKontext\citep{flux1kontext} & 0.347 & 0.126 & 15.4 & 0.701 & 37.8 & 17.3 & 0.514 \\
FluxKontext-MIE & \textcolor{red}{\textbf{0.627}} & \textcolor{red}{\textbf{0.468}} & \textcolor{red}{\textbf{25.7}} & \textcolor{red}{\textbf{0.916}} & \textcolor{red}{\textbf{51.3}} & \textcolor{red}{\textbf{23.1}} & \textcolor{red}{\textbf{0.726}} \\
\bottomrule
\end{tabular}
}
\end{table}

\subsection{Out-Of-Distribution Image Edit}

\begin{figure}[!h]
    \centering
    \includegraphics[width=\linewidth]{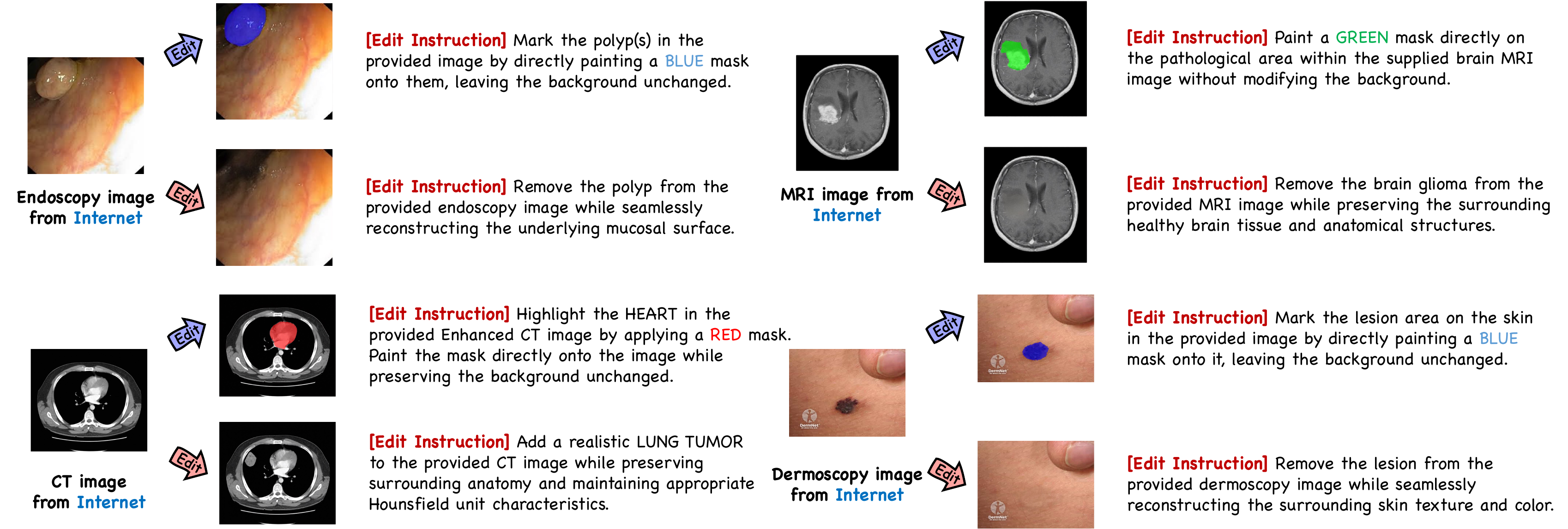}
    \caption{\textbf{Examples of Out-Of-Distribution Editing.}}
    \label{fig:ood}
\end{figure}

While Section~\ref{sec:generalization} demonstrates that the model trained on MieDB-100k generalizes effectively to OOD editing targets, we further evaluate its robustness by performing edits on `in-the-wild' medical images sourced from the internet (Fig.~\ref{fig:ood}).

The results indicate that our model is capable of readily adapting to medical images outside of datasets. This suggests that the diversity of MieDB-100k has successfully decoupled the model from specific data distribution, allowing it to internalize generalizable edit operations that are applicable to real-world clinical scenarios.

\subsection{Transformation Downstream Validation}
\label{app:downstream_mar}

To further demonstrate the practical clinical value of \datasetname, we conduct a downstream validation experiment on the metal artifact removal (MAR) task from the Transformation perspective. Specifically, we apply a pretrained all-organ segmentation model to both the original artifact-corrupted CT images and the restored CT images produced by editing models. By comparing the segmentation accuracy on the restored images against the ground-truth masks, we can quantitatively assess whether the editing model preserves clinically relevant anatomical structures after artifact removal.

\begin{table}[t]
\centering
\caption{\textbf{Downstream segmentation validation on metal artifact removal.} DICE scores between ground-truth masks and segmentation results on restored CT images.}
\label{tab:downstream_mar}
\begin{tabular}{l|cccc}
\toprule
\textbf{Restoration Model} & Qwen-Image-Edit & Nano Banana Pro & OmniGen2 & OmniGen2-MIE\\
\midrule
\textbf{DICE} & 0.472 & 0.699 & 0.355 & \textbf{0.873}\\
\bottomrule
\end{tabular}
\end{table}

As shown in Tab.~\ref{tab:downstream_mar}, OmniGen2-MIE significantly outperforming all baseline models. This result demonstrates that the model trained on \datasetname\ not only produces visually plausible restorations, but also faithfully preserves the underlying anatomical structures that are critical for downstream clinical analysis. This experiment further validates that \datasetname\ possesses significant real-world relevance and clinical value: training on our dataset enables editing models to produce outputs that genuinely useful for practical medical workflows such as computer-aided diagnosis on processed images.

\newpage
\subsection{Inpainting Quality Verification}
\subsubsection{Turing Test}
\label{app:turing_test}

To directly verify the quality of our FLUX-based healthy tissue inpainting, we conduct a Turing Test. Specifically, we collage 100 real healthy images and FLUX-generated healthy images side by side (with their left-right order randomized) and ask 3 medical experts to judge whether the image on the left is Good (more realistic), Same (comparable in quality), or Bad (more likely to be generated) relative to the image on the right.

\begin{table}[h]
\centering
\caption{\textbf{Turing Test results on FLUX-generated healthy images.} Good indicates the generated image fooled the expert; Same indicates comparable quality; Bad indicates the expert correctly identified the generated image.}
\label{tab:turing_test}
\begin{tabular}{l c c c}
\toprule
\textbf{Rating} & \textbf{Expert 1} & \textbf{Expert 2} & \textbf{Expert 3} \\
\midrule
Good (gen fooled) & 4\% & 8\% & 9\% \\
Same (equal quality) & 86\% & 77\% & 83\% \\
Bad (gen identified) & 10\% & 15\% & 8\% \\
\bottomrule
\end{tabular}
\end{table}

As shown in Tab.~\ref{tab:turing_test}, the vast majority of generated images were rated as Same or Good by all three experts, with the ``Same'' category consistently exceeding 77\%. This directly confirms that the inpainted healthy tissue achieves quality comparable to real medical images under expert clinical judgment, validating the fidelity of our Modification task data construction pipeline.

\subsubsection{Examples of Healthy Tissue Inpainting}

\begin{figure}[h!]
    \centering
    \includegraphics[width=0.7\linewidth]{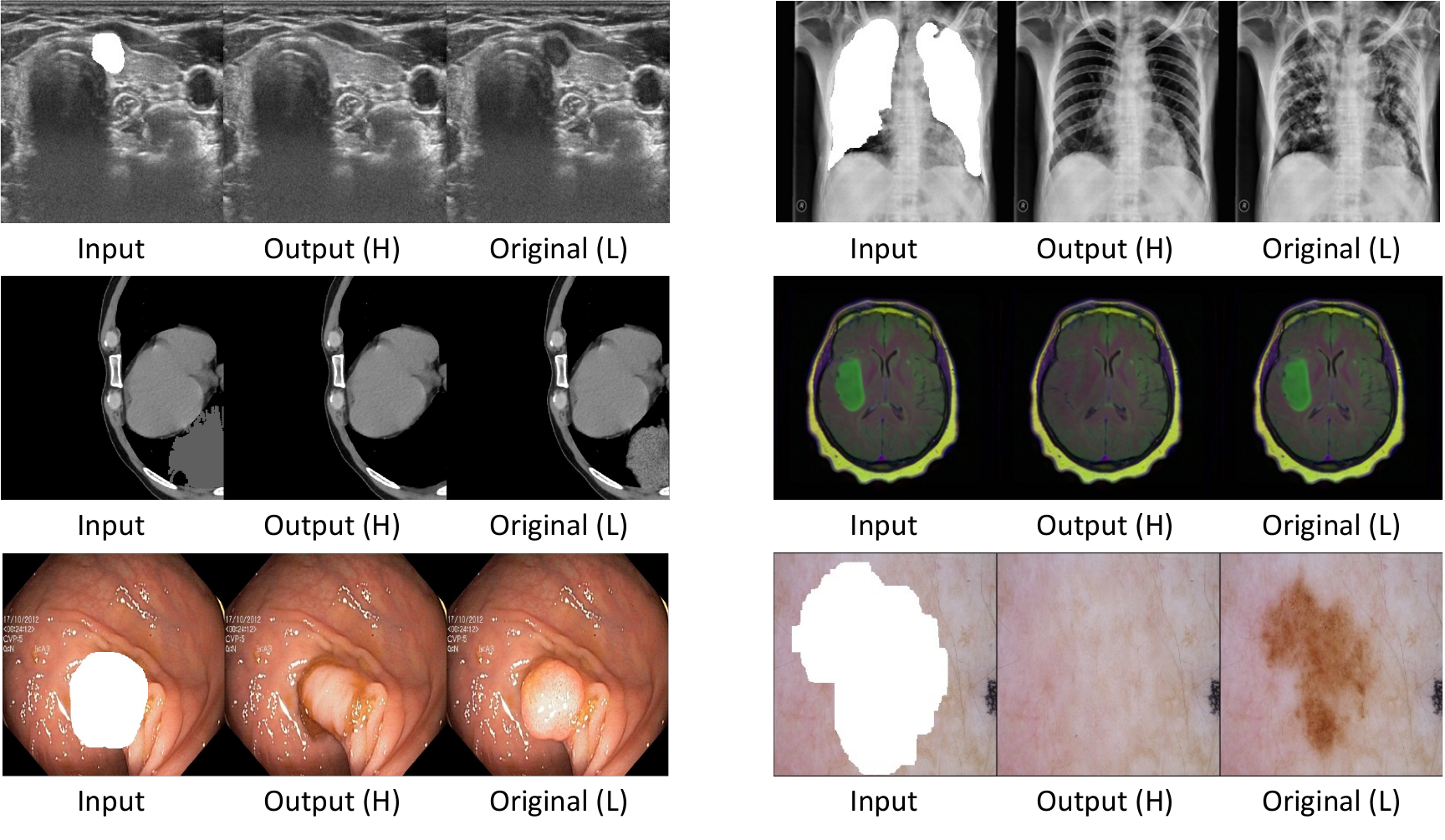}
    \caption{\textbf{Examples of Inpainting.} We train different inpainting models on each medical modalities. H: the Healthy image; L: the Lesion-bearing image.}
\end{figure}

\clearpage
\newpage
\section{More Qualitative Result}
\begin{figure}[!h]
    \centering
    \includegraphics[width=\linewidth]{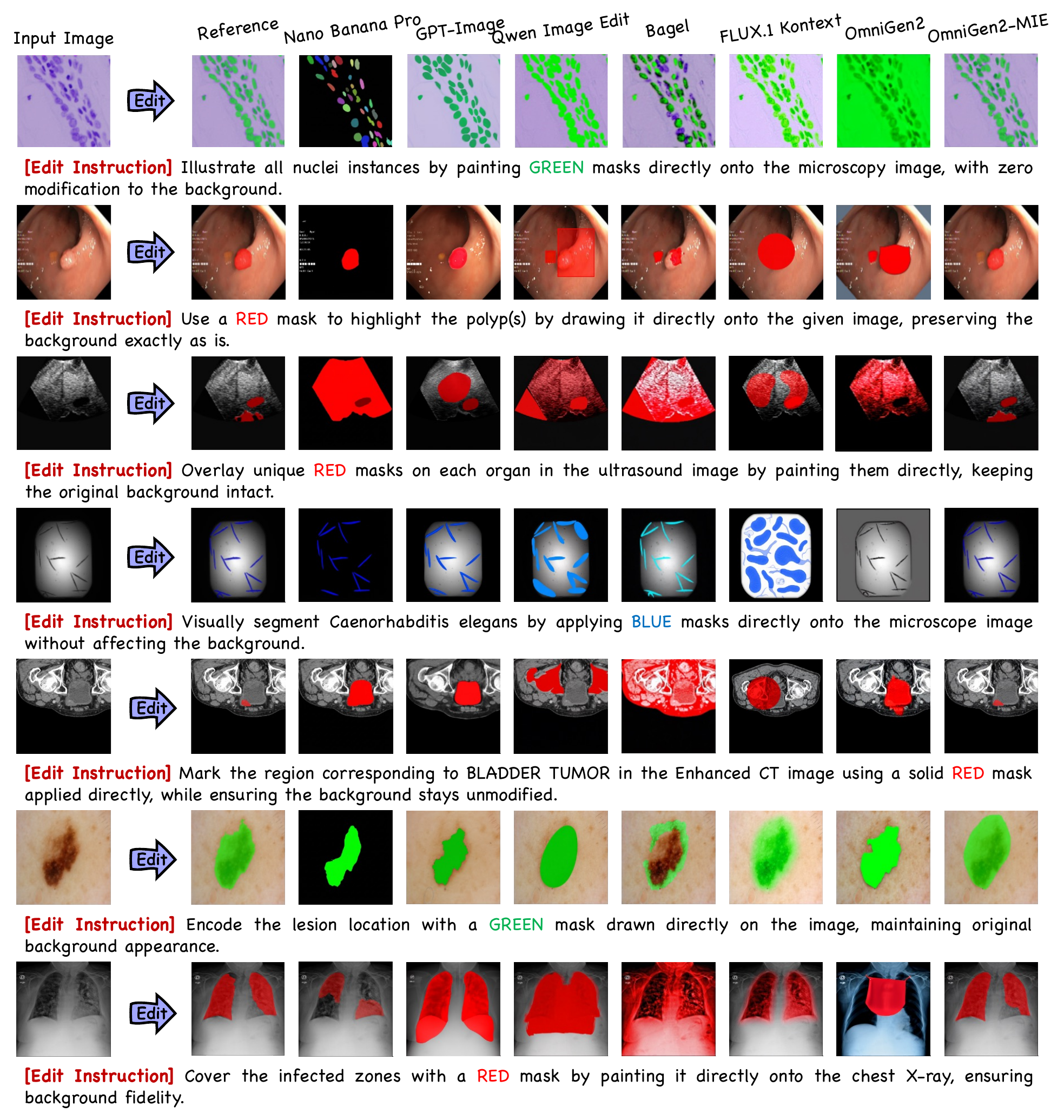}
\end{figure}

\begin{figure}[h]
    \centering
    \includegraphics[width=\linewidth]{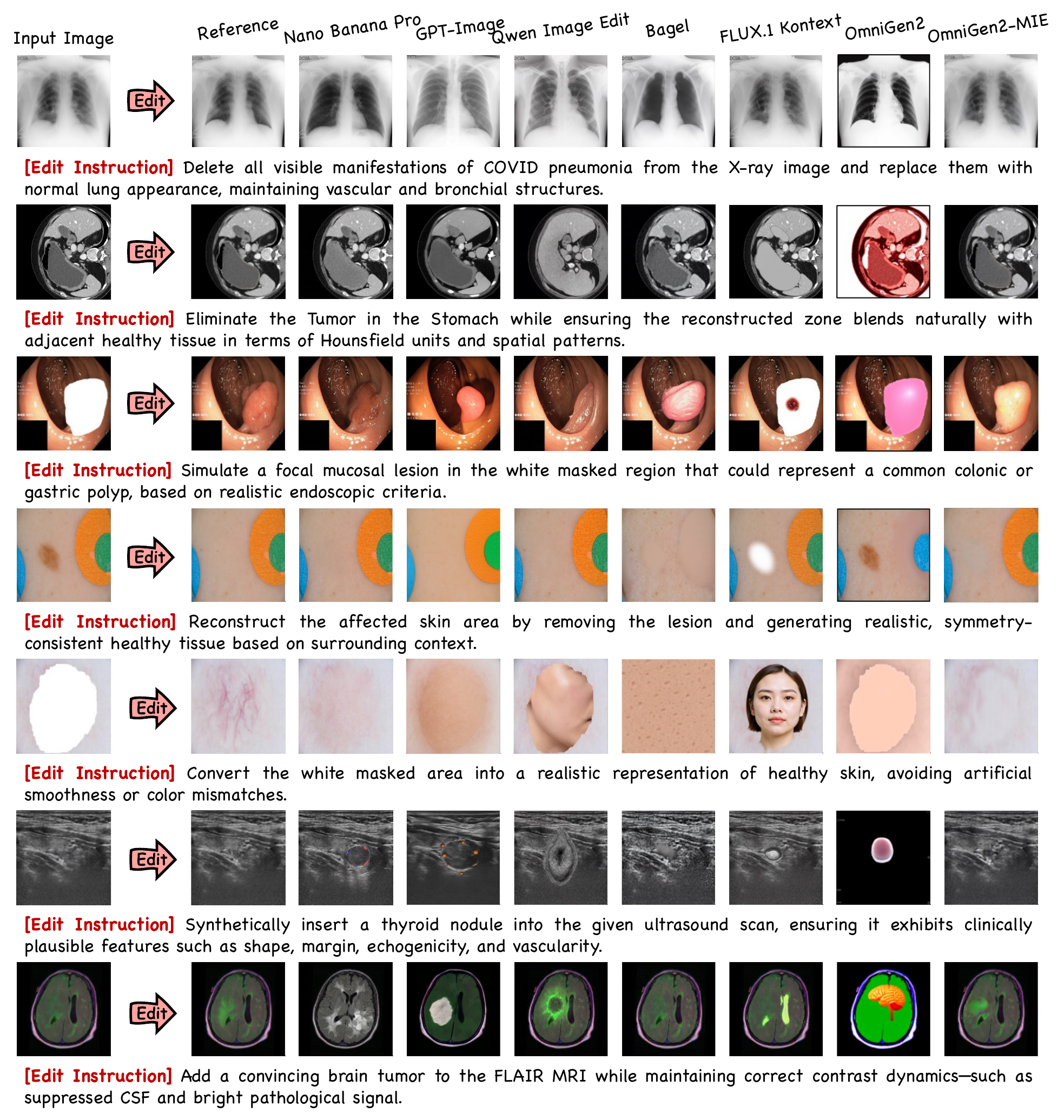}
\end{figure}

\begin{figure}[h]
    \centering
    \includegraphics[width=\linewidth]{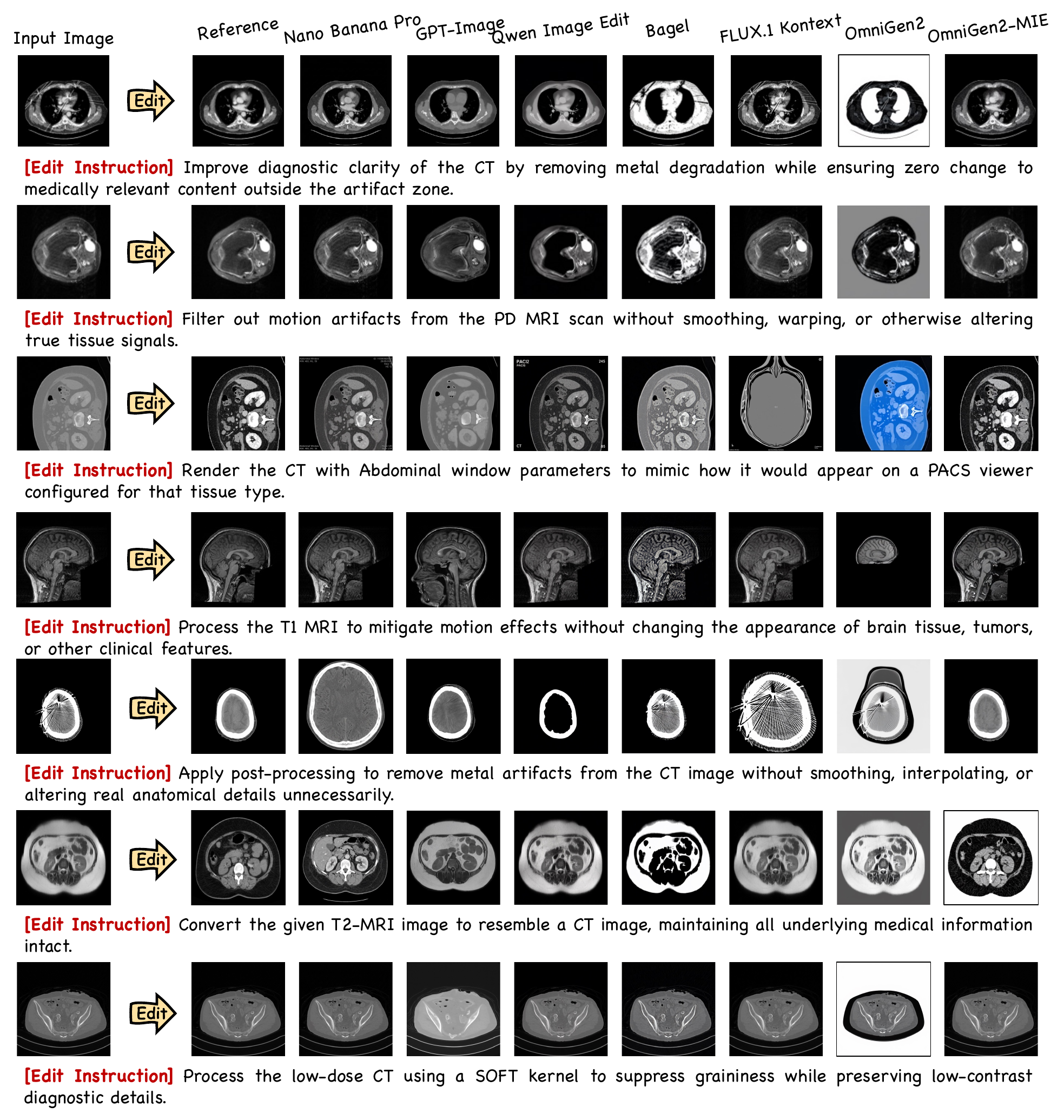}
\end{figure}

\clearpage
\newpage

\section{Failure Cases}
\label{app:failure_case}

\begin{figure}[!h]
    \centering
    \includegraphics[width=0.75\linewidth]{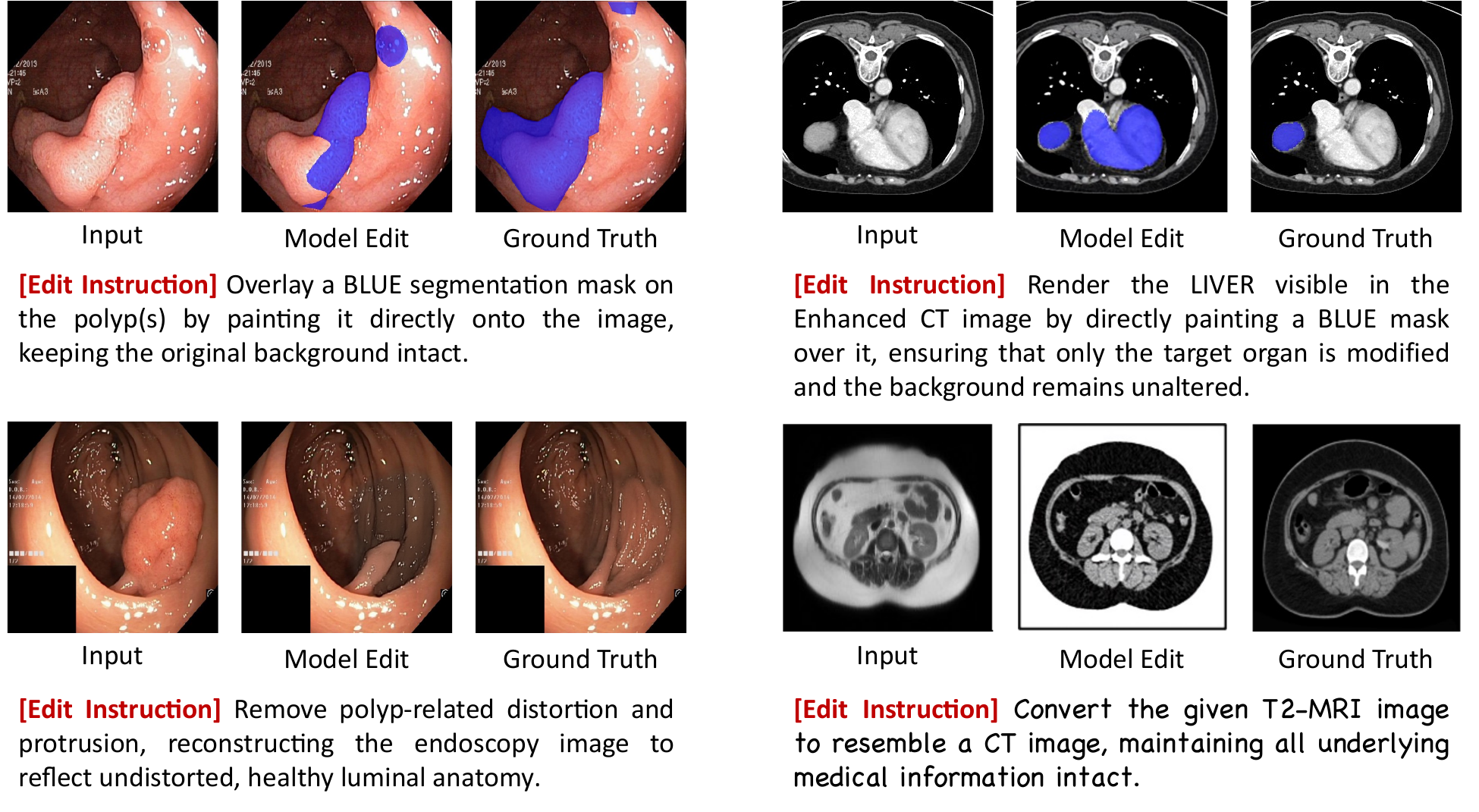}
    \caption{\textbf{Failure Cases.}}
    \label{fig:failure}
\end{figure}

Fig~\ref{fig:failure} illustrates several representative failure cases of OmniGen2-MIE. The most frequent failure modes include: (1) semantic confusion between targeted anatomical features and morphologically similar background tissues; (2) intensity inconsistency, where the brightness of the edited region deviates from the surrounding context in a physically implausible manner; and (3) background inconsistency, especially after holistic transformations. These limitations underscore the need for more sophisticated multimodal architectures capable of preserving fine-grained details, as well as even more comprehensive training datasets to satisfy the requirements of rigorous clinical applications.

\section{Limitations and Broader Impacts}
\label{app:limitations}

Despite MieDB-100k provides a large-scale and diverse dataset for medical image editing, the primary limitation lies in the inherent difficulty of capturing ALL possible medical imaging modalities, and the relative scarcity of data for rare clinical cases. Continuous efforts to enrich these underrepresented categories will be vital for enhancing the dataset's diversity and effectiveness. Furthermore, while our work establishes a foundation for unified understanding and generation, it focuses exclusively on editing tasks. Integrating medical VQA and text-to-image datasets represents a natural progression of this research direction, resulting in a more comprehensive resource for the development of holistic medical models.

Our work has two potential societal risks. 
First, generative medical image editing could be misused to fabricate or alter clinical evidence, posing risks to patient safety and diagnostic integrity; we therefore explicitly state that this dataset is intended solely for research purposes and is \textbf{not suitable for direct clinical decision-making}.
Second, biases present in the source datasets---such as under-representation of certain demographics, anatomical regions, or rare conditions---may propagate or be amplified by models trained on \datasetname, potentially leading to inequitable performance across patient populations.
We encourage downstream users to perform thorough bias audits and to comply with applicable regulatory and ethical guidelines before deploying any models trained on this data in clinical or real-world settings.

%% file: App/data_lisences.tex
\begin{table}[!h]
\centering
\caption{\textbf{Summary of public medical datasets utilized in the construction of MieDB-100k.} The columns \#Train and \#Benchmark denote the number of samples allocated to our training and benchmark splits respectively from each source dataset.}
\small
\resizebox{\linewidth}{!}{
\begin{tabular}{lcccc}
\toprule
\textbf{DatasetName} & \textbf{\#Train} & \textbf{\#Benchmark} & \textbf{Modality} & \textbf{License}\\ \midrule
AbdomenUS \citep{AbdomenUSMSBench} & 569 & 62 & Ultrasound & CC BY 4.0\\
Bbbc010 \citep{Bbbc010MSBench} & 70 & 20 & Microscopy & CC BY 4.0\\
Bkai-Igh \citep{Bkai-Igh-MSBench} & 700 & 81 & Endoscopy & CC BY 4.0\\
Brats-gli \citep{braintumorsegmentation} & 1529 & 80 & MRI & CC0 \\
BriFiSeg \citep{BriFiSegMSBench} & 1005 & 40 & Microscopy & CC BY 4.0\\
BUSI \citep{BusiMSBench} & 452 & 80 & Ultrasound & CC BY 4.0\\
CellNuclei \citep{CellNucleiMSBench} & 469 & 51 & Microscopy & CC BY 4.0\\
ChaseDB1 \citep{ChaseDB1MSBench} & 19 & 7 &  Fundus & CC BY 4.0\\ 
Chest-ct-segmentation \citep{Chest-CT-Segmentation} & 278 & 19 & CT & CC0\\
Chest-xray-masks-and-labels \citep{Chest-Xray-Masks-Labels} & 666 & 32 & Xray & CC0\\
CHUAC \citep{chuac} & 17 & 5 & Fundus & CC BY-NC 4.0\\ 
COVID-19\_Radiography\_Dataset \citep{Covid19RadioMSBench} & 2010 & 95 & Xray & MIT \\ 
COVID-19-CT-SCAN-Lesion \citep{MosMedPlusMSBench} & 255 & 15 & CT & CC BY 4.0\\
CovidQU \citep{CovidQUExMSBench} & 5684 & 122 & Xray & CC BY-SA 4.0\\
CT\_MAR \citep{ct-mar} & 1595 & 82 & CT & BSD 3-Clause\\
CT-Low-Dose-Reconstruction \citep{ct-low-dose} & 867 & 51 & CT & CC BY 4.0\\
CystoidFluid \citep{CystoFluidMSBench} & 703 & 59 & OCT & CC BY 4.0\\ 
Dca1 \citep{Dca1MSBench} & 93 & 28 & Fundus & CC BY-NC 4.0\\ 
Deepbacs \citep{DeepbacsMSBench} & 17 & 10 & Microscopy & CC BY 4.0\\
Drive \citep{DriveMSBench} & 18 & 20 & Fundus & CC BY 4.0\\ 
DynamicNuclear \citep{DynamicNuclearMSBench} & 50 & 17 & Microscopy & CC BY 4.0\\
FHPsAOP \citep{FHPsAOPMSBench} & 2800 & 80 & Ultrasound & CC BY 4.0\\
IDRiD \citep{IdribMSBench} & 47 & 27 & Fundus & CC BY 4.0\\  
ISIC2016 \citep{Isic2016MSBench} & 810 & 80 & Dermoscopy & CC0 \\
ISIC2018 \citep{Isic2018MSBench} & 9973 & 115 & Dermoscopy & CC0 \\
KMAR-50K \citep{kmar50k} & 651 & 47 & MRI & CC BY 4.0\\
Kvasir \citep{KvasirMSBench} & 4429 & 139 & Endoscopy & Custom (Research Only)\\
Lgg-mri-segmentation \citep{lgg-mri-seg} & 1669 & 55 & MRI & CC BY-NC-SA 4.0\\
MoNuSAC \citep{MonusacMSBench} & 0 & 21 & Microscopy & CC BY 4.0\\
MR-ART \citep{mr_art} & 820 & 18 & MRI & CC BY-NC 4.0\\
MSD \citep{msd} & 797 & 3912 & MRI & Apache-2.0 \\
NuSeT \citep{NusetMSBench} & 2383 & 40 & Microscopy & CC BY 4.0\\
Paired\_MRI\_CT \citep{Paired_ct_mri} & 1974 & 72 & CT, MRI & Custom (Research Only)\\
Pandental \citep{PandentalMSBench} & 81 & 24 & Xray & CC BY 4.0\\
Pasta-GEN \citep{pasta-gen} & 32299 & 731 & CT & MIT\\
PolypGen \citep{PolypGenMSBench} & 984 & 75 & Endoscopy & CC BY 4.0\\
PROMISE12 \citep{Promise12MSBench} & 1031 & 80 & MRI & CC BY 4.0\\
QaTa-COV19 \citep{qatacov} & 3573 & 85 & Xray & CC BY-NC-SA 3.0\\
Refuge \citep{refuge} & 80 & 80 & Fundus & CC BY 4.0\\ 
RoboTool \citep{RoboToolMSBench} & 350 & 76 & Surgical Photo & CC BY 4.0\\
ThyroidXL \citep{thyroidxl} & 7029 & 138 & Ultrasound & Custom (Research Only)\\
Tnbcnuclei \citep{TnbcnucleiMSBench} & 35 & 10 & Microscopy & CC BY 4.0\\
TotalSegmentator \citep{total_seg} & 5206 & 154 & CT, MRI & CC BY 4.0\\
UltrasoundNerve \citep{UltrasoundNerveMSBench} & 1651 & 50 & Ultrasound & CC BY 4.0\\
USforKidney \citep{USforKidneyMSBench} & 4351 & 50 & Ultrasound & CC BY 4.0\\
UWSkinCancer \citep{UWSkinCancerMSBench} & 143 & 44 & Dermoscopy & CC BY 4.0\\
VinDr-Multiphase \citep{Vindr} & 3486 & 44 & CT & CC BY 4.0\\
WBC \citep{WbcMSBench} & 280 & 40 & Microscopy & CC BY 4.0\\
YeaZ \citep{YeazMSBench} & 358 & 51 & Microscopy & CC BY 4.0\\
YGA\_low\_dose\_ct \citep{YGA} & 4387 & 44 & CT & CC BY 4.0\\
\bottomrule
\end{tabular}
}
\end{table}

%% file: App/vlm_filter.tex
\begin{tcolorbox}[
    enhanced,
    breakable,
    colback=blue!3,
    colframe=blue!40!black,
    colbacktitle=blue!60!black,
    coltitle=white,
    fonttitle=\bfseries\large,
    title=VLM Filter Prompt for Endoscopy,
    arc=6pt,
    boxrule=0.5pt,
    left=15pt,
    right=15pt,
    top=10pt,
    bottom=15pt,
    after skip=1cm
]
\begin{small}
\begin{verbatim}
You are a specialized Medical AI Evaluator for endoscopic image
inpainting. Your task is to assess the quality of an AI-edited
image where a polyp has been removed and replaced with synthetic
tissue.

### Task
Evaluate the center frame (Edited Image) by comparing it with
the right frame (Original Image) and the left frame (Inpainting
Mask). You must determine if the polyp removal is medically
plausible or if the AI has introduced "unnatural content."
For clarity, the polyp area in the center frame has also been
highlighted by red bounding box.

### Focus Area
Concentrate your evaluation ONLY on the region indicated by the
WHITE mask in the leftmost frame.

### Criteria Definitions
- **image_quality**:
    - "good": Sharp mucosal texture, proper lighting, clear
      vascular patterns.
    - "borderline": Mild blur, slight overexposure/underexposure,
      but still interpretable.
    - "poor": Severe motion blur, pixelation, or lighting that
      obscures anatomy.
- **has_polyp_or_unnatural_content**:
    - "obvious": Presence of residual polyp, or medically
      impossible structures like an abrupt hole (perforation-like),
      a "crater," or a "donut" protrusion.
    - "minor": Unnatural flatness or slight bulge that doesn't
      look like healthy mucosa.
    - "no": The area looks like smooth, healthy, and continuous
      intestinal mucosa.

### Strict Output Format
Return ONLY a JSON object. Do not include any conversational
filler.
{
  "image_quality": "good" | "borderline" | "poor",
  "has_polyp_or_unnatural_content": "obvious" | "minor" | "no",
  "reasoning": "A concise explanation (max 2 sentences) focusing
      on why the content is considered unnatural or successful."
}

You are not making a medical diagnosis, only screening image
quality and obvious abnormalities.
Return ONLY a JSON object, no additional text or explanation.
\end{verbatim}
\end{small}
\end{tcolorbox}

%% file: App/rubric.tex
\begin{tcolorbox}[
    enhanced,
    breakable,
    colback=pink!5,       
    colframe=pink!30,     
    colbacktitle=red!50, 
    coltitle=white,       
    fonttitle=\bfseries\large,
    title=Scoring Rubric for Modification Tasks,
    arc=6pt,              
    boxrule=0.5pt,        
    left=15pt,            
    right=15pt,
    top=10pt,
    bottom=15pt,
    after skip=1cm
]
You are a helpful assistant in evaluating medical image editing result. 

You will be provided with an edit instruction and a collage image where the leftmost is origin image, center is edited image and rightmost is the reference ground truth image.

You should score how well an edited image matches the intended edit while preserving clinical realism and image integrity based on following scoring rubrics:

    \begin{enumerate}[label=\protect\textbf{\arabic*)}, leftmargin=*, itemsep=1em]
        \item \textbf{Edit Goal Fulfillment (Edit Correctness)}: Assesses whether the intended lesion change is achieved.

        Scoring reference:
            \begin{itemize}[label=--, nosep]
                \item \textit{5: Lesion added/removed exactly as intended; no residuals or unintended remnants.}
                \item \textit{4: Mostly correct; slight residual signal after removal or slight under/over-addition.}
                \item \textit{3: Partial success; lesion still partially present (removal) or incomplete/incorrect lesion (addition).}
                \item \textit{2: Wrong area or wrong type of change; target lesion largely unchanged.}
                \item \textit{1: No effective edit or opposite edit performed.}
            \end{itemize}

        \item \textbf{Edit Area Morphology (Shape, Margins, Internal Structure)}: 
        Evaluates whether edit area matches expected morphology and/or reference.

        Scoring reference:
            \begin{itemize}[label=--, nosep]
                \item \textit{5: Shape, border characteristics, and internal texture are highly consistent.}
                \item \textit{4: Minor border/shape irregularities; still plausible.}
                \item \textit{3: Morphology is generic/unconvincing; borders/texture inconsistent.}
                \item \textit{2: Clearly artificial morphology (blocky, repeated patterns, unnatural contours).}
                \item \textit{1: Morphology nonsensical or misleading (e.g., appears like different pathology).}
            \end{itemize}

        \item \textbf{Intensity / Signal / Attenuation Consistency}: Checks whether edited region match modality-specific intensities.

        Scoring reference:
            \begin{itemize}[label=--, nosep]
                \item \textit{5: Intensities match local tissue statistics; no intensity discontinuities.}
                \item \textit{4: Slight intensity mismatch detectable with careful viewing.}
                \item \textit{3: Obvious mismatch (too bright/dark), inconsistent with modality or anatomy.}
                \item \textit{2: Strong intensity discontinuity; clearly edited.}
                \item \textit{1: Severe intensity errors that invalidate the image (e.g., saturation/clipping, inverted contrast).}
            \end{itemize}
            
        \item \textbf{Boundary Blending \& Transition Naturalness}: Rates edge blending and transitions between edited and unedited regions.
        
        Scoring reference:
            \begin{itemize}[label=--, nosep]
                \item \textit{5: Seamless blending; no halos, ringing, cut-paste edges.}
                \item \textit{4: Minor halo/edge artifacts only on close inspection.}
                \item \textit{3: Visible seams; boundary looks edited.}
                \item \textit{2: Strong cutout appearance or blur patches.}
                \item \textit{1: Boundary artifacts dominate the image.}
            \end{itemize}

        \item \textbf{Background / Non-target Preservation}: Measures unintended changes outside the lesion edit region.
        Scoring reference:
            \begin{itemize}[label=--, nosep]
                \item \textit{5: Non-target anatomy and background unchanged (within expected noise).}
                \item \textit{4: Small unintended changes but not clinically meaningful.}
                \item \textit{3: Noticeable unintended alterations in nearby structures.}
                \item \textit{2: Large unintended modifications to anatomy or overall image.}
                \item \textit{1: Global corruption or major anatomical distortions.}
            \end{itemize}

        \item \textbf{Anatomical Plausibility \& Clinical Coherence}: Assesses whether result respects anatomy and pathology logic (e.g., lesion doesn't cross impossible boundaries).
        Scoring reference:
            \begin{itemize}[label=--, nosep]
                \item \textit{5: Fully plausible; consistent with organ boundaries and expected presentation.}
                \item \textit{4: Mostly plausible; minor oddity but acceptable.}
                \item \textit{3: Questionable plausibility (e.g., lesion overlaps structures unnaturally).}
                \item \textit{2: Clearly implausible anatomy/pathology relationship.}
                \item \textit{1: Clinically nonsensical or misleading.}
            \end{itemize}

        \item \textbf{Artifact Introduction (Noise, Texture, Aliasing, Compression, Repetition)}: Evaluates new artifacts introduced by editing.
        Scoring reference:
            \begin{itemize}[label=--, nosep]
                \item \textit{5: No new artifacts; noise texture consistent with original.}
                \item \textit{4: Minor artifacts (subtle smoothing/grain mismatch).}
                \item \textit{3: Artifacts visible and distracting.}
                \item \textit{2: Strong artifacts (banding, checkerboard, repeated texture).}
                \item \textit{1: Severe artifacts preventing clinical use.}
            \end{itemize}
        
        \item \textbf{Image Quality \& Acquisition Consistency}: Checks consistency with scanner characteristics (resolution, blur, point spread, slice thickness cues, motion).
        Scoring reference:
            \begin{itemize}[label=--, nosep]
                \item \textit{5: Matches acquisition characteristics; sharpness/noise consistent.}
                \item \textit{4: Slight mismatch in sharpness or noise level.}
                \item \textit{3: Clear mismatch (over-smoothed or over-sharpened region).}
                \item \textit{2: Strong mismatch; edited region appears from different source.}
                \item \textit{1: Completely inconsistent with acquisition; unusable.}
            \end{itemize}
            
    \end{enumerate}

    \vspace{0.5em}
    \textbf{Return Format:} Return a JSON dictionary with two fields:
    \begin{itemize}[label=\tiny$\blacksquare$]
        \item ``\texttt{conclusion}'': A brief conclusion to the edited image.
        \item ``\texttt{score\_list}'': The scores of the eight aspects in a JSON list.
    \end{itemize}

\end{tcolorbox}

%% file: checklist.tex
\section*{NeurIPS Paper Checklist}

\begin{enumerate}

\item {\bf Claims}
    \item[] Question: Do the main claims made in the abstract and introduction accurately reflect the paper's contributions and scope?
    \item[] Answer: \answerYes{}
    \item[] Justification: The abstract and introduction clearly state three contributions: (1) proposing MieDB-100k with a scalable curation pipeline (Section~\ref{sec:method}), (2) unifying medical image understanding and generation into editing (Section~\ref{sec:pipeline}), and (3) comprehensive benchmarking demonstrating training effectiveness (Section~\ref{sec:experiment}). All claims are supported by experimental results.

\item {\bf Limitations}
    \item[] Question: Does the paper discuss the limitations of the work performed by the authors?
    \item[] Answer: \answerYes{}
    \item[] Justification: A Limitations section is provided in Appendix~\ref{app:limitations}.

\item {\bf Theory assumptions and proofs}
    \item[] Question: For each theoretical result, does the paper provide the full set of assumptions and a complete (and correct) proof?
    \item[] Answer: \answerNA{}
    \item[] Justification: This paper presents a dataset and empirical benchmarking study. The only mathematical derivation (alpha de-blending for mask reconstruction in Appendix~\ref{app:mask_restore}) is fully detailed with equations and assumptions.

    \item {\bf Experimental result reproducibility}
    \item[] Question: Does the paper fully disclose all the information needed to reproduce the main experimental results of the paper to the extent that it affects the main claims and/or conclusions of the paper (regardless of whether the code and data are provided or not)?
    \item[] Answer: \answerYes{}
    \item[] Justification: The data construction pipeline is fully described in Section~\ref{sec:pipeline}. All training hyperparameters for OmniGen2-MIE are provided in Appendix~\ref{app:train_detail}. Evaluation metrics and protocols are detailed in Section~\ref{sec:evaluation} and Appendix~\ref{app:eval_detail}. All 48+ source datasets are listed in Appendix~\ref{app:data_list}. The dataset and code are publicly released.

\item {\bf Open access to data and code}
    \item[] Question: Does the paper provide open access to the data and code, with sufficient instructions to faithfully reproduce the main experimental results, as described in supplemental material?
    \item[] Answer: \answerYes{}
    \item[] Justification: The dataset and code are publicly available at \url{https://anonymous.4open.science/r/MieDB-100k-3BB5}, as stated in the abstract.

\item {\bf Experimental setting/details}
    \item[] Question: Does the paper specify all the training and test details (e.g., data splits, hyperparameters, how they were chosen, type of optimizer) necessary to understand the results?
    \item[] Answer: \answerYes{}
    \item[] Justification: Training details including learning rate, batch size, number of steps, precision, and scheduler are provided in Section~\ref{sec:experiment} and Appendix~\ref{app:train_detail}. Data splits are described in Section~\ref{sec:post_process}, with strict separation of source dataset train/test splits to prevent data leakage. Baseline model configurations follow their official inference settings.

\item {\bf Experiment statistical significance}
    \item[] Question: Does the paper report error bars suitably and correctly defined or other appropriate information about the statistical significance of the experiments?
    \item[] Answer: \answerYes{}
    \item[] Justification: We report Pass@3 results in Appendix~\ref{app:multi_round} to account for generative variance. Inter-rater reliability is validated via pairwise Spearman correlations ($\rho > 0.957$, all $p < 1\times10^{-5}$) in Section~\ref{sec:evaluation}. Cross-evaluator consistency ($\rho = 0.951$, $p = 2\times10^{-5}$) is reported in Appendix~\ref{app:cross_evaluator}. The correlation between human preference and VLM scoring ($\rho = -0.905$, $p = 3\times10^{-4}$) is also provided.

\item {\bf Experiments compute resources}
    \item[] Question: For each experiment, does the paper provide sufficient information on the computer resources (type of compute workers, memory, time of execution) needed to reproduce the experiments?
    \item[] Answer: \answerYes{}
    \item[] Justification: The details of GPUs and training configuration (20,000 steps, batch size 64, BF16 precision) are specified in Appendix~\ref{app:train_detail}. 
    
\item {\bf Code of ethics}
    \item[] Question: Does the research conducted in the paper conform, in every respect, with the NeurIPS Code of Ethics \url{https://neurips.cc/public/EthicsGuidelines}?
    \item[] Answer: \answerYes{}
    \item[] Justification: The research uses only publicly available, de-identified medical image datasets. All source datasets have undergone their own institutional de-identification and ethical review processes. The paper discusses limitations and potential risks in Appendix~\ref{app:limitations}.

\item {\bf Broader impacts}
    \item[] Question: Does the paper discuss both potential positive societal impacts and negative societal impacts of the work performed?
    \item[] Answer: \answerYes{}
    \item[] Justification: Positive impacts are discussed in Introduction, Section~\ref{sec:perception} and Appendix~\ref{app:downstream_mar} (assisted interpretation, medical education, clinical screening support). Negative impacts and risks are discussed in Appendix~\ref{app:limitations}. The dataset is explicitly stated as not suitable for direct clinical decision-making.
    
\item {\bf Safeguards}
    \item[] Question: Does the paper describe safeguards that have been put in place for responsible release of data or models that have a high risk for misuse (e.g., pre-trained language models, image generators, or scraped datasets)?
    \item[] Answer: \answerYes{}
    \item[] Justification: The dataset is constructed exclusively from publicly available, de-identified medical image datasets, minimizing privacy risks. The dataset is intended for research purposes only. All source datasets have undergone their respective institutional review processes, and the paper clearly states the dataset should not be used for direct clinical decision-making.

\item {\bf Licenses for existing assets}
    \item[] Question: Are the creators or original owners of assets (e.g., code, data, models), used in the paper, properly credited and are the license and terms of use explicitly mentioned and properly respected?
    \item[] Answer: \answerYes{}
    \item[] Justification: All 48+ source datasets are listed in Appendix~\ref{app:data_list} under research-free license. All baseline models used in experiments are cited in Section~\ref{sec:experiment}.

\item {\bf New assets}
    \item[] Question: Are new assets introduced in the paper well documented and is the documentation provided alongside the assets?
    \item[] Answer: \answerYes{}
    \item[] Justification: MieDB-100k is thoroughly documented: data definition, construction pipeline, source datasets, evaluation metrics, and limitations. The dataset and code are released with documentation at the GitHub repository.

\item {\bf Crowdsourcing and research with human subjects}
    \item[] Question: For crowdsourcing experiments and research with human subjects, does the paper include the full text of instructions given to participants and screenshots, if applicable, as well as details about compensation (if any)? 
    \item[] Answer: \answerYes{}
    \item[] Justification: The paper involves three board-certified medical experts for data quality control and evaluation. The evaluation protocol, including task-specific descriptions, representative examples, and the Good/Fair/Poor classification criteria, is detailed in Appendix~\ref{app:manual_detail}. Human preference ranking criteria are described in Section~\ref{sec:evaluation}.

\item {\bf Institutional review board (IRB) approvals or equivalent for research with human subjects}
    \item[] Question: Does the paper describe potential risks incurred by study participants, whether such risks were disclosed to the subjects, and whether Institutional Review Board (IRB) approvals (or an equivalent approval/review based on the requirements of your country or institution) were obtained?
    \item[] Answer: \answerNA{}
    \item[] Justification: The human involvement in this work does not constitute human subjects research. The medical experts participated as professional annotators assessing data quality, not as study subjects. No patient data was directly collected.

\item {\bf Declaration of LLM usage}
    \item[] Question: Does the paper describe the usage of LLMs if it is an important, original, or non-standard component of the core methods in this research? Note that if the LLM is used only for writing, editing, or formatting purposes and does \emph{not} impact the core methodology, scientific rigor, or originality of the research, declaration is not required.
    \item[] Answer: \answerYes{}
    \item[] Justification: LLMs/VLMs are used as core components in the data construction and evaluation pipeline: Qwen3-VL-32B-Instruct for rejection sampling in Modification tasks, Qwen-Max for prompt rephrasing, and GPT-5.2 as the automated evaluator for rubric scoring. All usages are explicitly described.

\end{enumerate}